\RequirePackage{fix-cm}
\documentclass[twocolumn]{svjour3}           %
\smartqed  %
\usepackage{graphicx}
\usepackage{url}              %
\usepackage{booktabs}         %
\usepackage{amsfonts} 
\usepackage{amsmath}
\usepackage{amssymb}%
\usepackage{color}            %
\usepackage{xcolor}    %
\usepackage{multirow}
\usepackage{adjustbox}
\usepackage{caption} %
\usepackage[misc]{ifsym}
\usepackage[pagebackref=true,breaklinks=true,letterpaper=true,colorlinks,bookmarks=false]{hyperref}
\usepackage[labelfont=bf]{caption}
\usepackage{colortbl}
\usepackage{tabularx}
\usepackage{tikz,dcolumn,subcaption}
\usepackage{comment}
\definecolor{turquoise}{cmyk}{0.65,0,0.1,0.3}
\definecolor{purple}{rgb}{0.65,0,0.65}
\definecolor{dark_green}{rgb}{0, 0.5, 0}
\definecolor{orange}{rgb}{0.8, 0.6, 0.2}
\definecolor{red}{rgb}{0.8, 0.2, 0.2}
\definecolor{darkred}{rgb}{0.6, 0.1, 0.05}
\definecolor{blueish}{rgb}{0.0, 0.3, .6}
\definecolor{light_gray}{rgb}{0.8, 0.8, 0.8}
\definecolor{pink}{rgb}{1, 0, 1}
\definecolor{greyblue}{rgb}{0.25, 0.25, 1}
\definecolor{mistyrose}{rgb}{1.0, 0.89, 0.88}
\definecolor{whitee}{rgb}{1.0, 1.0, 1.0}
\definecolor{palerobineggblue}{rgb}{0.59, 0.87, 0.82}
\definecolor{lavenderblue}{rgb}{0.9, 0.9, 1.0}

\definecolor{darkblue}{HTML}{6082B6}
\definecolor{darkorange}{HTML}{FF8C00}

\newcommand{\liu}[1]{{\color{black}#1}}

\begin{document}

\title{FunnyNet-W: Multimodal Learning of Funny Moments in Videos in the Wild}

\author{Zhi-Song Liu$^{1}$, Robin Courant$^{2}$ Vicky Kalogeiton$^{2}$}
\authorrunning{Liu et al.}

\institute{
    $^1$ Computer Vision and Pattern Recognition Laboratory, Lappeenranta-Lahti University of Technology, Finland \\
    $^2$LIX, Ecole Polytechnique, IP Paris \\
}

\date{Received: XXX / Accepted: XXX}

\maketitle

\abstract{
Automatically understanding funny moments (i.e., the moments that make people laugh) when watching comedy is challenging, as they relate to various features, such as body language, dialogues and culture. 
In this paper, we propose FunnyNet-W, a model that relies on cross- and self-attention for visual, audio and text data to predict funny moments in videos. Unlike most methods that rely on ground truth data in the form of subtitles, in this work we exploit modalities that come naturally with videos: (a) video frames as they contain visual information indispensable for scene understanding, (b) audio as it contains higher-level cues associated with funny moments, such as intonation, pitch and pauses and (c) text automatically extracted with a speech-to-text model as it can provide rich information when processed by a Large Language Model.  
To acquire labels for training, we propose an unsupervised approach that spots and labels funny audio moments. 
We provide experiments on five datasets: the sitcoms TBBT, MHD, MUStARD, Friends, and the TED talk UR-Funny. 
Extensive experiments and analysis show that FunnyNet-W successfully exploits visual, auditory and textual cues to identify funny moments, while our findings reveal FunnyNet-W's ability to predict funny moments in the wild. FunnyNet-W sets the new state of the art for funny moment detection with multimodal cues on all datasets with and without using ground truth information.

\keywords{Multimodal learning \and Vision+language \and Video understanding \and Humor detection}

}

\maketitle

\section{Introduction}
\begin{figure}[t]
	\centering
		\centerline{\includegraphics[width=\columnwidth]{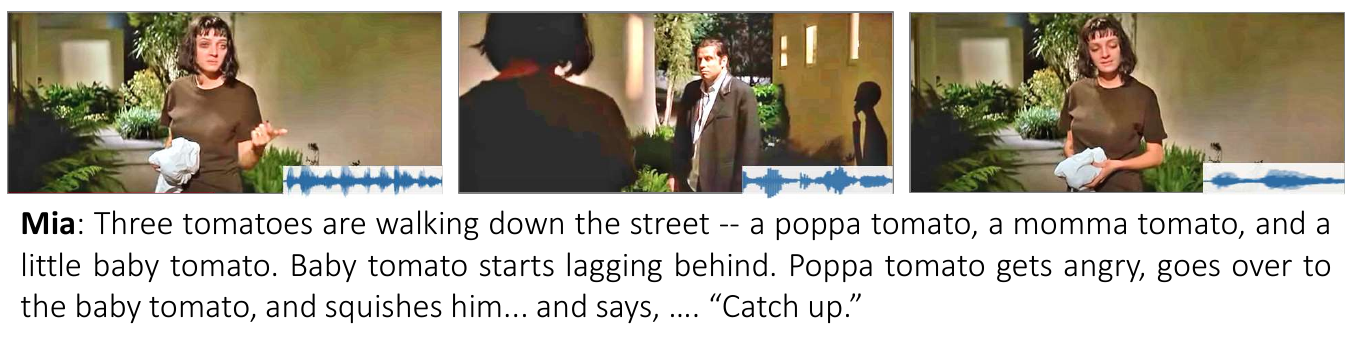}} 
		\caption{
        What is funny? Audio cues along with visual frames and textual data are a rich source of information for identifying funny moments in videos.  Video scene from Pulp Fiction, 1994, source video \url{https://www.youtube.com/watch?v=4L5LjjYVsHQ}
		}
		\label{fig:teaser}
\end{figure}

We understand the world by using our senses, especially in multimedia areas. All signals can stimulate one's feelings and reactions. 
Funniness is universal and timeless: in 1900 BC Sumerians wrote the first joke and it is still funny nowadays.
However, whereas humans can easily understand funny moments, even from different cultures and eras, machines do not.
Even though the number of interactions between humans and machines is growing fast, identifying funniness is still a brake on making these interactions spontaneous.
Actually, understanding funny moments is a complex concept since they can be purely visual, purely auditory, or they can mix both cues: there is no recipe for the perfect joke. 

Recently, there have been attempts to understand the nature of jokes, humour, and funny moments~\cite{colbert,rjokes}. However, most of these works have relied solely on textual cues, with only a few incorporating videos~\cite{humor_data,laught_machine}. 
The limitation of these approaches lies in their dependence on external transcripts in the form of manual subtitles, which are not naturally available with raw video data.
In contrast, advancements in the field of speech-to-text have made it easier to extract accurate transcripts from raw audio waveforms that naturally accompany videos. This enables processing natural language to better understand the overall context. 
Furthermore, including audio as a modality in the funny moment detection pipeline is essential, as raw audio carries essential and complementary cues, including tones, pauses, pitch, pronunciation, and background noises~\cite{accent,mustard}. When speaking, the way people convey their message is as important as the actual content being delivered. 
Similarly, visual content plays a crucial role. For example, the same phrase spoken by the same person can elicit different emotional responses depending on the context (see Figure~\ref{fig:teaser}). Facial expressions, body gestures, and scene context contribute to a better understanding of the intended meaning, thereby influencing the perceived funniness.

Therefore, in this paper, we introduce FunnyNet-W, a multimodal model for predicting funny moments in videos. It comprises three encoders: (a) visual encoder, which captures the global contextual information of a scene; (b) textual encoder, which represents the overall understanding of a scene; and (c) audio encoder, which captures voice and language effects; and the Cross Attention Fusion (CAF) module, i.e., a new module that learns cross-modality correlations hierarchically so that features from different modalities can be combined to form a unified feature for prediction. Thus, FunnyNet-W is trained to learn to embed all cross-attention features in the same space via self-supervised contrastive learning~\cite{simclr}, in addition to classifying clips as funny or not funny.
To obtain labeled data, we exploit the laughter that naturally exists in sitcom TV shows. We define as `funny-moment' any $n$-second clip followed by laughter; and `not-funny' the clips not followed by laughter. 
To extract laughter, we propose an unsupervised labeling approach that clusters audio segments into laughter, music, voice and empty, based on their waveform difference\footnote{Note that we use the laughter solely as an indicator for data labeling, but the laughter is not the included in the audio segments of FunnyNet-W. Once FunnyNet-W is trained, it can detect funny moments in any video, with or without laughter.}. Moreover, we enrich the Friends dataset with laughter annotations.

Our extensive experimentation and analysis show that combining audio, visual and textual cues (that all come naturally with videos) is suitable for funny-moment detection. 
Moreover, we compare FunnyNet-W to the state of the art on five datasets including sitcoms (TBBT, MHD, MUStARD, and Friends) and TED talks (UR-Funny), and show that it outperforms all other methods for all metrics and input configurations. Note that even by using only automatically generated text from audio, FunnyNet-W outperforms all other methods that rely on ground-truth text in the form of subtitles.  
Furthermore, we examine the difference between our proposed FunnyNet-W and automatic chatbots based on Large-Language-Models (LLMs). Our findings show that without specific prompt engineering under the few-shot setting, chatbots cannot understand the funniness of texts. Instead, our proposed FunnyNet-W significantly outperforms chatbots in prediction accuracy, highlighting the importance of specific multimodal training for this task. 
We also apply FunnyNet-W to data from other domains, i.e., movies, stand-up comedies, and audiobooks. For quantitative evaluation, we apply FunnyNet-W on a sitcom without canned laughter manually annotated. It shows that FunnyNet-W predicts funny moments without fine-tuning, revealing its flexibility for funny-moment detection in the wild.

Our contributions are summarized as follows: 
(1) We introduce FunnyNet-W, a model for funny moment detection that uses audio, visual, and textual modalities that come automatically with videos. FunnyNet-W combines features from the three modalities using the proposed CAF module relying on cross and self-attention; 
(2) Extensive experiments and analysis highlight that FunnyNet-W successfully exploits audio, visual and textual cues;
(3) FunnyNet-W achieves the new state of the art on five datasets. We also demonstrate its generalizability by comparing it to automatic LLM chatbots and its flexibility by showcasing in-the-wild applications. 
The code is available online on the project page: {\small{\url{https://www.lix.polytechnique.fr/vista/projects/2024_ijcv_liu/}}}. 

A preliminary version of this work has been published in
ACCV 2022~\cite{funnynet}. We significantly extend it in the following ways:
\begin{itemize}
    \item  \textbf{Motivation.} We propose FunnyNet-W, a multimodal model for funny moment detection in videos. FunnyNet-W follows the same motivation from FunnyNet, i.e. leverage modalities that come with videos for free. Given that most funny moments are inherently associated with language, in addition to the audiovisual features of FunnyNet, FunnyNet-W leverages speech-to-text features. For this, we automatically generate text from speech by leveraging Automatic Speech Recognition methods, and then pair it with the rich representation capability of Large Language Models (LLMs), thus enabling to better understand the specificities of language. This is motivated thoroughly in the Introduction, in Section~\ref{sub:differences} and experimentally evaluated in the new Sections~\ref{subsub:encoders}, \ref{subsub:modalities}, \ref{sub:analysis_modality}, and \ref{sub:analysis_visualization}. 
    \item \textbf{Architecture.} FunnyNet uses audio, visual and face encoders to process the multimodal signals. The face encoder, however, is cumbersome and requires an external face detection model. For this reason, in FunnyNet-W we do not use a face encoder. Instead, FunnyNet-W uses an LLM text encoder to process textual data that are automatically transcribed. Moreover, in FunnyNet-W, we use a more modern visual encoder. The differences between the two models are described in Section~\ref{sub:differences} and experimentally compared in Sections~\ref{sub:sota} (Table~\ref{tab:sota1}) and \ref{subsub:encoders}. 
    \item \textbf{Experiments and analysis.} We provide more insights and content to explain the performance of FunnyNet-W.  
    Specifically, we experimentally demonstrate and discuss the benefits of the new encoders, each modality and their fusion module, of the length of the input time window, of the losses used as opposed to alternative ones (Section~\ref{sub:ablation}). Furthermore, we provide a thorough qualitative and intuitive analysis of each modality and their fusion, as well as failure cases (Section~\ref{sec:analysis}). 
    \item \textbf{In the wild applications.} In addition to experimenting on other domains as in FunnyNet (Section~\ref{sub:coolappli}), we perform two in-the-wild applications: first, we compare FunnyNet-W against chatbots based on LLMs and show that relying solely on language with or without prompt engineering is insufficient for detecting funnyness  (Section~\ref{sub:wild_chatbot}); and second, we replace real speech by synthetic speech and showcase the importance of real vocals for funny moment detection (Section~\ref{sub:wild_audio}).  
\end{itemize}

\section{Related Work}
\label{sec:relwork}
\paragraph{Sarcasm and Humor Detection.}
Sarcasm and humor share similar styles (irony, exaggeration and twist) but also differ from each other in terms of representation. 
Sarcasm usually relates to dialogues; hence, most methods detect sarcasm by processing language using human efforts. For instance, \cite{sarcasm_1} collects a speech dataset from social media using the hashtag and manual labeling, while others~\cite{sarcasm_2,sarcasm_3} study the acoustic patterns related to sarcasm, like slower speaking rates or higher volumes of voice. 
In contrast, a humorous moment is defined as the moment before laughter~\cite{mustard,urfunny}. Hence, such methods~\cite{audio_humor,mustard,urfunny,laught_machine,hasan2021humor} process audios to extract laughter for labeling. 
Nevertheless, for prediction, most such approaches focus solely on language models~\cite{colbert,rjokes} or on multiple cues including text~\cite{urfunny,hasan2021humor}. 
For instance, LaughMachine~\cite{laught_machine} proposes vision and language attention mechanisms, while MSAM~\cite{humor_data} combines self-attention blocks and LSTMs to encode vision and text. 
\cite{MISA} use first an advanced BERT~\cite{bert} model to process long-term textual correlation and then vision for the prediction. Following this, \cite{mag_transformer} propose a Multimodal Adaptation Gate to efficiently leverage textual cues to explore better representation for sentiment analysis. OxfordTVG-HIC~\cite{li2023oxfordtvghic} proposes a dataset with 2.9 M image-text pair for humor detection. 
A few methods also explore audio. For instance, MUStARD~\cite{mustard} and URFUNNY~\cite{urfunny} process text, audio and frames using LSTM to explore long-term correlations, while HKT~\cite{hasan2021humor} classifies language (context and punchline) and non-verbal cues (audio and frame) to learn cross-attention correlations for humor prediction. They combine audio with other information (video and texts) in a simple feature fusion process without investigating the inter-correlations in depth. Specifically, they stack multimodal features to learn the global weighting parameters without considering the biases in different domains. 
In contrast, we believe that funny scenes can be triggered by mutual signals from multimodalities; hence, in this work, we explore the cross-domain agreement of cues with contrastive training. 
Moreover, FunnyNet-W eliminates the need for external textual annotation by relying solely on raw audiovisual cues, and extracts textual cues directly from the audio that naturally accompanies videos.

\paragraph{Sound Event Detection and Laughter detection.} 

\noindent \textit{Sound event detection} aims to identify and timestamp sound events within audio recordings. 
Most attempts either rely on annotated data~\cite{mesaros2016tut} or use source separation techniques~\cite{demucs,demucs_last}. 
The choice of input representation is crucial, and most methods use Mel spectrograms~\cite{mesaros2017detection,wang2021multimodal,byol_a,byola_v2,cola} instead of audio waveforms. This choice is motivated by their computational efficiency, interpretability, and effortless integration into conventional vision models.
In our work, we focus on a specific acoustic event: laughter. We leverage these detected laughter as pseudo-labels to train FunnyNet-W. 

\noindent \textit{Laughter detection.} The literature in this domain remains relatively scarce. Some methods rely on physiological sensors~\cite{barral2017no,laughlog}, while others~\cite{old_laughter_detector,new_laughter_detector} follow the conventional supervised learning paradigm to train deep neural laughter detectors.
Nevertheless, the latter approach requires annotated datasets, a challenging endeavour in the context of this specialized domain.  
For instance, the authors of~\cite{new_laughter_detector} experiment with the Switchboard dataset~\cite{switchboard}, which contains manually annotated laughter timestamps from phone conversation, and also manually annotate laughter timestamps from 1000 clips of AudioSet dataset~\cite{gemmeke2017audio}. 
In contrast, our laughter detector is unsupervised, robust and straightforward, by leveraging the specific attributes of multichannel audio data. Our method sidesteps the need for complex annotations, presenting a promising alternative within the laughter detection landscape.

\paragraph{Multimodal tasks.}
Over the past decade,  the number of tasks that require multiple modalities has increased either due to their intrinsic multimodal nature or due to the potential performance enhancements of adding extra modalities. Here, we review some approaches that are directly related to our work in terms of multimodality and modality fusion.

\textit{Audio+Video.} For instance,~\cite{audio_vision_1,audio_vision_2} recognize the facial movements to separate the speaker's voice in the audio. 
\cite{audio_vision_3,audio_vision_4} temporally align the audio and video using 
attention to locate the speaker. The former~\cite{audio_vision_3} proposes a triplet network to process the query, positive and negative samples to encourage the query to be close with positive samples and far from negative samples. The latter~\cite{audio_vision_4} collects an Audio-Visual Event (AVE) dataset to better handle audio-visual alignment. Several methods extend this to other applications, such as audiovisual generation~\cite{audio_vision_10} that generates an audio-driven talking face from a single source image and pose video.

\textit{Video+Language.} Several tasks involve combining language and visual--in particular video-- modalities.
One notable category encompasses video-to-text tasks, including video captioning \cite{wang2020event,lin2022swinbert}, which entails generating natural language captions for video sequences. A more challenging, yet very similar task is video question answering \cite{liang2020learning,yang2023vid2seq,zhu2022end}, where the goal is to comprehend the content well enough to respond to queries effectively.
In contrast, Singer et al.~\cite{singer2022make} propose an approach focusing on text-to-video generation.
Finally, video-text retrieval~\cite{dong2021dual,bain2021frozen,fang2021clip2video} aims to facilitate bidirectional exploration of both video and textual content.

\textit{Audio+Language.} Numerous research directions focus only on audiovisual modalities. 
A major audiovisual task lies in speech emotion recognition~\cite{yoon2018multimodal,priyasad2020attention}, which aims to connect audio and text to categorize emotions. A speech emotion recognition pipeline consists of modality fusion followed by classification. 
Parallel to the well-established image-text retrieval task, the domain of audio-text retrieval has also received substantial attention~\cite{lou2022audio,koepke2022audio,xin2023improving,mei2022language}. This task employs similar techniques based on measuring feature similarity between the modalities. 
Another complex audiovisual challenge is audio captioning, where the objective is to generate textual descriptions from acoustic inputs. Most approaches rely on the classical encoder-decoder architecture~\cite{koizumi2020transformer,shen2023fine,kim2023prefix}.

\textit{Video+Language+Audio.}
Some works have extended previous tasks by combining the three modalities: audio, video and text.
For example, certain approaches incorporate the acoustic modality into the conventional video captioning pipeline \cite{iashin2020multi,liu2022visually}. Additionally, \cite{deng2018multimodal} introduce an acoustic modality to enhance emotion recognition. \liu{\cite{avlnet} learns a shared audio-visual embedding space directly from raw video inputs via self-supervision. \cite{autoad} use CLIP~\cite{clip} to align audio-visual signals to produce audio descriptions. \cite{hyperbolic} propose a hyperbolic loss to align audio-visual features in a tree-shaped space.} All these works show improvements in comparison to unimodal baselines. 
In contrast to previous methods that depend on distinct annotated sources and ground-truth modalities (for instance subtitles for text or ground-truth annotations), our proposed FunnyNet-W extracts multiple additional modalities-- audio and text-- from a single modality source, namely video, using non-perfect extraction techniques, such as speech-to-text models.

\paragraph{Modality alignment.}
Recently many works~\cite{clip,guzhov2022audioclip,girdhar2023imagebind} have shown promising efforts for acquiring shared multimodal embeddings by leveraging large-scale datasets. 
Notably, the first breakthrough of text and image embedding was achieved with CLIP~\cite{clip}. 
Comparable milestones have been reached in diverse modalities; for instance, \cite{morgado2021audio} proposes to learn a powerful audio-visual representation from videos. 
Other works extend the original CLIP language-image representation with new modalities such as audio in\cite{guzhov2022audioclip}, or video in~\cite{lin2022frozen,xue2023clip}. Recently, the ImageBind~\cite{girdhar2023imagebind} unifies six distinct modalities into a shared embedding space.
The key to success consists in aligning features from the different modalities.

Attention mechanisms~\cite{transformer} is natural for connecting multimodal signals. For instance, \cite{cross_attn_2} employ cross-attention to model inter- and intra-modality relationships, \cite{tanCOMMA2021} leverage contrastive cross-attention, \cite{jaegle2021perceiver,lee2020cross}  use iterative cross-attention. \liu{In addition, Nagrani et al. \cite{nagrani2021attention} introduced attention bottlenecks with randomly initialized bottleneck tokens for modality fusion. In contrast, our fusion mechanism builds on this idea but differs by (i) employing modality projections as bottlenecks and (ii) integrating an extra self-attention block to capture fused token correlations}. 
These works illustrate the natural strength of attention mechanisms in aligning multiple modalities within a unified space.  
Numerous multimodal tasks benefit from this capability of modality alignment. 
Applications such as summarization~\cite{narasimhan2021clip}, retrieval~\cite{gabeur2020multi,bain2021frozen}, audiovisual classification~\cite{nagrani2021attention}, 
predicting goals~\cite{epstein2021learning}, human replacement~\cite{dufour22eccv}. \cite{tanCOMMA2021,cross_attn_2} iteratively apply self and cross-attention to explore correlations among modalities. 
Instead, FunnyNet-W both fuses all modalities and in parallel learns the cross-correlation among different modalities; this avoids any biases that may be caused by one dominant modality.

\section{Method}
\label{sec:method}
Here, we present FunnyNet-W, its training process and losses (Sections~\ref{sub:archi}-\ref{sub:training_losses}). 
For training labels, we propose an unsupervised laughter detector (Section~\ref{sub:laughter_det}). 

\begin{figure*}[t]
\centering
		\centerline{\includegraphics[width=\linewidth]{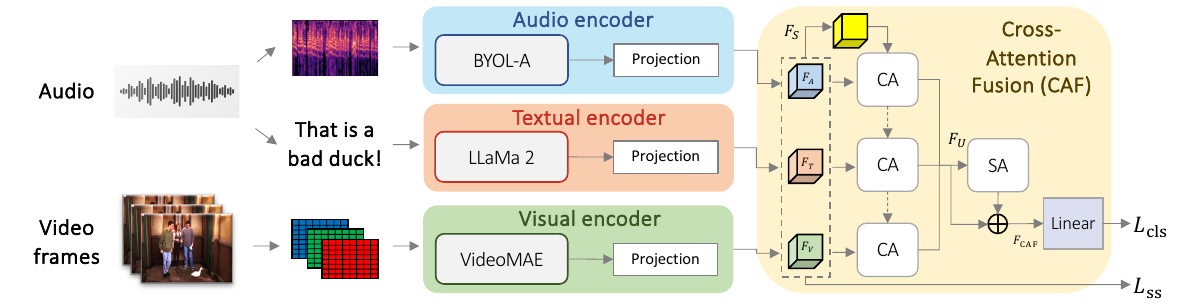}}
		\caption{
        Architecture of FunnyNet-W.
		Given audio-visual clips, FunnyNet-W predicts funny moments in videos. It consists of the audio (\textit{blue}), textual (\textit{red}), and visual (\textit{green}) encoders, whose outputs pass through the Cross Attention Fusion (\textit{CAF}), which consists of cross-attention (\textit{CA}) and self-attention (\textit{SA}) for feature fusion. It is trained to embed all modalities in the same space via self-supervision ($L_{\text{ss}}$) and to classify clips as funny or not-funny ($L_{\text{cls}}$). 
		}
		\label{fig:main}
\end{figure*}

\noindent \textbf{Overview.} FunnyNet-W consists of 
(i) three encoders: the visual encoder with videos as input, the audio encoder with audio as input, and the text encoder with subtitles as input. To parse the subtitles, we use an automatic speech recognition (ASR) system~\cite{radford2022whisper}; 
(ii) the proposed Cross-Attention Fusion (CAF) module,  which explores cross- and intra-modality correlations by using cross- and self-attentions in the encoders' outputs. 
Then, the fused feature is fed to a binary classifier. 
The overall architecture is illustrated in Figure~\ref{fig:main}.
FunnyNet-W is trained to embed all modalities in the same space via self-supervised contrastive loss and to classify clips as funny or not.  
For training, we exploit laughter that naturally exists in TV Shows: we define it as `funny-moment' for any audiovisual snippet followed by laughter; and `not-funny' for any audiovisual snippet not followed by laughter.

\subsection{FunnyNet-W Architecture}
\label{sub:archi}

FunnyNet-W utilizes raw inputs from videos, including the audio waveform and frames.

\noindent \textbf{Audio Encoder.}
First, the audio waveform is transformed into a Mel spectrogram\footnote{Mel spectrogram is a 2D acoustic time-frequency representation of sound.}. This spectrogram, denoted as $\mathbf{X}_{\text{audio}}$, is then passed through an audio encoder to generate a 1D feature vector. Finally, a projection head is applied to obtain a $N$-dimensional vector $\mathbf{F_{\text{A}}}\in \mathbb{R}^{N}$.

\noindent \textbf{Text Encoder.}
The corresponding transcripts, denoted as $X_{\text{text}}$, are extracted from the audio waveform using an automatic speech recognition model~\cite{radford2022whisper}. 
These transcripts are then encoded into a feature vector using the text encoder. Subsequently, a projection is performed to obtain a $N$-dimensional vector $\mathbf{F_{\text{T}}}\in \mathbb{R}^{N}$.

\noindent \textbf{Visual Encoder.} 
The visual encoder employs an architecture based on the transformer to process video frames. The input frames, denoted as $\mathbf{X}_{\text{visual}}$, are divided into patches from several consecutive frames. Unlike conventional approaches that use a `classification token' to obtain a general representation, we compute the representation by averaging the pooled features from all patches. This process results in a feature vector, which is then projected to a $N$-dimensional vector $\mathbf{F}_{\text{V}}\in \mathbb{R}^{N}$ using a projection head. The video context complements the audio in providing richer content~\cite{urfunny}. Additionally, in the absence of sound and transcripts, visual cues can also elicit laughter.

\noindent \textbf{Projection Head.} 
This module consists of two linear layers separated by a GeLU activation function. Dropout and normalization layers are applied after the linear layers. It takes the features outputted from each encoder and projects them into a shared $N$-dimensional multimodal feature space.

\noindent \textbf{Cross-Attention Fusion (CAF).} It learns the cross-domain correlations among vision, audio and text (yellow box Figure~\ref{fig:main}). It consists of (a) three cross-attention (CA) and (b) one self-attention (SA) modules, described below:
\begin{itemize}
    \item     \textbf{Cross-attention} is used in cross-domain knowledge transfer to learn across-cue correlations by attending the features from one domain to another~\cite{cross_attn_1,cross_attn_3,cross_attn_2}. 
In CAF, it models the relationship among vision, audio, and textual features. 
We stack all features as $\mathbf{F_{\text{S}}}{\in} \mathbb{R}^{3\times512}$, and then feed $\mathbf{F_{\text{S}}}$ into three cross-attention modules to attend to \liu{vision, text, and audio}, respectively (Figure~\ref{fig:main}). 
Next, the scaled attention per modality is computed as $\sigma\left(\frac{\mathbf{Q_{\text{S}}}\mathbf{K_{i}^T}}{\sqrt{d}}\right)\mathbf{V_{i}}$, where $i{=}\{V,T,A\}$ for \{vision, text,  audio\}, and  $\sigma$ the softmax. 
The query $\mathbf{Q}$ comes from the stacked features:  $\mathbf{Q_{\text{S}}} {=} \mathbf{F_{\text{S}}}\mathbf{W}^{\text{Q}_{\text{S}}}$, while the key $\mathbf{K}$ and value $\mathbf{V}$ come from a single modality as $\mathbf{K_{i}} {=}\mathbf{F_{i}} \mathbf{W}^{\text{K}_{i}}$, and $\mathbf{V_{i}} {=}\mathbf{F_{i}} \mathbf{W}^{\text{V}_{i}}$.
Next, we obtain three cross-attentions and sum them to a unified feature $\mathbf{F_{\text{U}}}$ as: 

\begin{equation} 
\text{F}_{\text{U}} = \sum_{i \in \{V,F,A\}} \sigma\left(\frac{\mathbf{Q_{\text{S}}}\mathbf{K_{i}^T}}{\sqrt{d}}\right)\mathbf{V_{i}} \quad .
\label{eq:ca_attn1}
\end{equation}

 \item \textbf{Self-attention} computes the intra-correlation of the $F_{\text{U}}$ features, which are further summed with a residual $F_{\text{U}}$ as:

\begin{equation} 
\text{F}_{\text{CAF}} = \mathbf{F_{\text{U}}} + \sigma\left(\frac{\mathbf{Q_{\text{U}}}\mathbf{K_{\text{U}}^{T}}}{\sqrt{d}}\right)\mathbf{V_{\text{U}}} \quad ,
\label{eq:sa_attn}
\end{equation}

\noindent where $\mathbf{Q_{\text{U}}}{=}\mathbf{F_{\text{U}}} \mathbf{W}^{\text{Q}_{\text{U}}}$, $\mathbf{K_{\text{U}}}{=}\mathbf{F_{U}}\mathbf{W}^{\text{K}_{\text{U}}}$, $\mathbf{V_{\text{U}}}{=}\mathbf{F_{U}} \mathbf{W}^{\text{V}_{\text{U}}}$. 
Finally, we average $F_{\text{CAF}}$ tokens and feed it to a classification layer.
\end{itemize}

\paragraph{Discussion.} CAF differs to existing methods~\cite{cross_attn_1,cross_attn_2} in the computation of the cross attention. 
Using stacked features $F_{\text{S}}$ to attend to each modality $Q_{\text{S}}$ brings three benefits: 
(a) it is order-agnostic: for any  modality pair we compute cross-attention once, instead of twice by interchanging queries and keys/values; this results in reduced computation;  
(b) each modality serves as a query to search for tokens in other modalities; this brings rich feature fusion; and (c) it generalizes to any number of modalities, resulting in scalability\footnote{\liu{Note that although the CAF module scales linearly with the number of modalities, the total training time complexity is increased quadratically with the number of modalities ($\mathcal{O}(d^2)$), because all modality pairs are taken into consideration when computing the loss (see Equation~\ref{eq:selfsup}). However, not all loss pairs are necessary if one modality plays a dominant role, then we can skip some loss pairs. Our experiments in Table~\ref{tab:ablation} also show the importance of different modalities. For instance, the increase in F1 from the single text-only $T$ model to the $T+A$ model is 7\%, while the increase of F1 from the $T+A$ model to the $T+A+V$ model is 3.7\%. This indicates that it may not be necessary to use more modalities if the information they offer is overlapping and not complementary.}}.

\subsection{Training Process and Loss Functions}
\label{sub:training_losses}

\textbf{Subtitle extraction.} To extract transcripts from the raw waveform, we use the WhisperX system~\cite{bain2022whisperx}. 
WhisperX enforces alignment of the automatic speech recognition model Whisper \cite{radford2022whisper} with an external voice activity detection model to produce accurate word-level timestamps. This approach results in a Time-Accurate Speech Transcription, very similar to manually transcribed subtitles.

\noindent \textbf{Positive and Negative Samples.} 
To create samples, we exploit the laughter that naturally exists in episodes. 
We define as `funny' any $n$-sec clip followed by laughter; `not-funny' any $n$-sec clip not followed by laughter. 
More formally, given a laughter at timestep $(t_s, t_e)$, we extract a $n$-sec clip at $(t_s{-}n, t_s)$ and we split it into audio and video. For each video, we sample $n$ frames (1 FPS). 
For the audio, we resample it at 16000 Hz and transform it to Mel spectrogram. 
Thus, each sample corresponds to $n$ sec and consists of a Mel spectrogram for the audio and a $n$-frame long video. In practice, we use 8-sec clips as the average time between two canned laughters, and it also leads to better performances (ablations of $n$-sec clips and $n$-frames per clip in supplementary). Note that we clip the audio based on the starting time of the laughter so the positive samples do not include any laughter.

\noindent \textbf{Self-Supervised Contrastive Loss.} To capture `mutual' audiovisual information, we solve a self-supervised synchronization task~\cite{chung2016out,korbar2018co,owens2018audio}: we encourage visual features to be correlated with true audios and uncorrelated with audios from other videos.  
Given the i-th pair of visual $v^i$ and true audio features  $a^{i}$ and $N$ other audios from the same batch: $a_{1}, ..., a_{N}$ we minimize the loss~\cite{simclr,chung2019perfect,oord2018representation}:

\begin{equation} 
L_{\text{cotrs}} = - \text{log} \frac{\text{exp}(\text{S}(v^i, a^i) / \tau) }{\sum_{j=1}^{N} \text{exp}(\text{S}(v^i, a^j)/\tau) } \quad , 
\label{eq:contra}
\end{equation}

\noindent where $\text{S}$ the cosine similarity and $\tau$ the temperature factor. 
Equation~\ref{eq:contra} accounts for audio and visual features. 
Here, we compute the contrastive loss between all three modalities, i.e., visual-audio, text-audio, and visual-text. 
Thus, our self-supervised loss is:

	\begin{equation} 
L_{\text{ss}} = - \frac{1}{3} \big(L_{\text{cotrs}}^{v^{i},a^{i}} + L_{\text{cotrs}}^{v^{i},t^{i}} + L_{\text{cotrs}}^{t^{i},a^{i}} \big).
\label{eq:selfsup}
	\end{equation}

\noindent \textbf{Final Loss.}
FunnyNet is trained with a Softmax loss $Y_{\text{cls}}$ to predict if the input is funny or not, and the $L_{\text{ss}}$ to learn `mutual' information across modalities. 
Thus, the final loss is: 
\begin{equation} 
L {=} \lambda_{\text{ss}} L_{\text{ss}} + \lambda_{\text{cls}} L_{\text{cls}} \quad, 
\end{equation} 

\noindent where $\lambda_{\text{ss}}$, $\lambda_{\text{cls}}$ the weighting parameters that control the importance of each loss.

\subsection{Unsupervised Laughter Detection}
\label{sub:laughter_det}
\begin{figure*}[t]
	\centering
\centerline{\includegraphics[width=\linewidth]
  {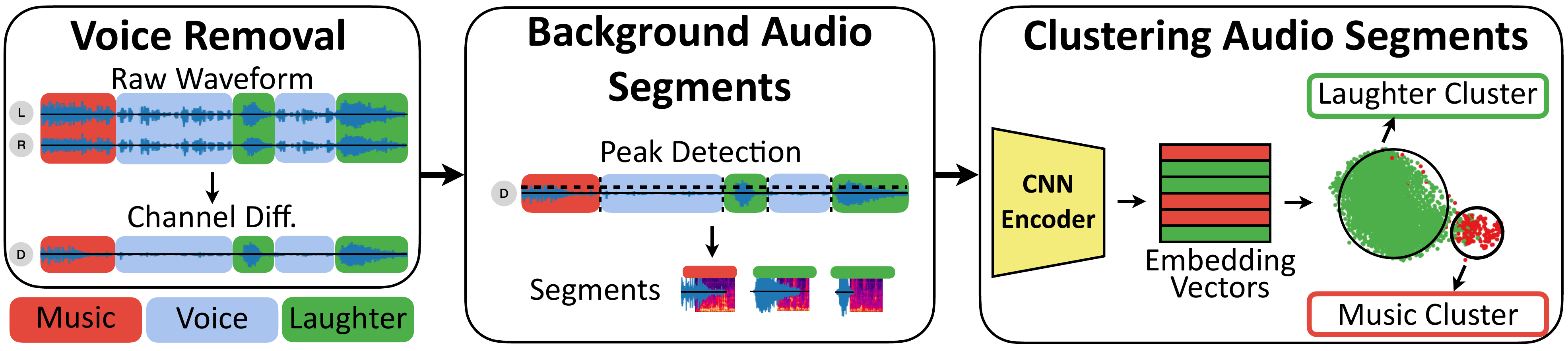}}
		\caption{
		\small{
        Proposed laughter detector.
		It takes raw waveforms as input and consists of (i) removing voices by subtracting channels (here, the audio is stereo with 2 channels), (ii) detecting peaks, and (iii) clustering audios to music and laughter. 
		}} 
		\label{fig:audio_pipeline}
\end{figure*}

To detect funny moments, we design an unsupervised laughter detector consisting of 3 steps (Figure~\ref{fig:audio_pipeline}). 

\begin{enumerate} 
\item \textbf{Remove Voices.}
Background audios include sounds, music, laughter; instead, voice (speech) is part of the foreground audio. 
We remove voices from audios by exploiting multichannel audio specificities. 
Given raw waveform audios, when the audio is stereo (two channels), the voices are centered and are common in both channels~\cite{huber2012modern}; hence, by subtracting the channels, we remove the voice and keep the background audio. 
In surround tracks (six channels), we remove the voice channel~\cite{huber2012modern} and keep the background ones.
\item \textbf{Background Audios.}
The waveforms from (i) are mostly empty with sparse peaks corresponding to audio: laughter and music. 
To split them into background and empty segments, we use an energy-based peak detector\footnote{\url{https://github.com/amsehili/auditok}.} that detects peaks based on the waveform energy. 
Then, we keep background segments and convert them to log-scaled Mel spectrograms. 
\item \textbf{Cluster Audio Segments.}
For each laughter and music segment, we extract features using a self supervised pre-trained encoder. 
Then, we cluster all audio segments using K-means to distinguish the laughter from the music ones.

\end{enumerate}

\subsection{Differences to the ACCV 2022 version~\cite{funnynet}}
\label{sub:differences}

FunnyNet~\cite{funnynet} and FunnyNet-W use multimodal input signals from videos and identify whether a video input is funny or not. Here, we describe the three main methodological differences between the two models. 

FunnyNet~\cite{funnynet} 
uses three encoders: (a) visual encoder for video frames (Timesformer~\cite{timesformer}) (b) audio encoder for voice and background audio (BYOL~\cite{byol_a}) and (c) face encoder for facial expressions (ResNet~\cite{facenet}). However, the face encoder is cumbersome as it requires an external model for face detection, thus leading to higher runtime and decreasing its applicability. 
For this reason, in FunnyNet-W we remove the face encoder, which leads to slight performance drops but increased gains in applicability and scalability. 
Moreover, FunnyNet-W uses the more modern visual encoder VideoMAE. 

Furthermore, for FunnyNet-W, we made the following three-fold observation. First, most funny moments are inevitably related to language. Second, recent advances in Automatic Speech Recognition (ASR)~\cite{radford2022whisper,bain2022whisperx} have rendered it possible to exploit the vocal part of the audio (i.e., the part where people are speaking) and automatically transcribe existing dialogues. 
Third, the recent explosion of Large Language Models (LLMs) offers remarkable capabilities in processing text and dialogues across a wide range of tasks. 
Combining these would mean transcribing dialogues via ASR \emph{for free}, then using an LLM encoder to detect funniness. However, FunnyNet does not rely on textual data. We consider this a wasted opportunity, as the textual data via LLMs can boost the representation capability of the model and, in turn, its performance. To this end, FunnyNet-W differs from FunnyNet in two aspects: it relies on ASR to transcribe dialogues \emph{for free} and it uses an LLM text encoder for processing the text (Llama-2~\cite{llama2}).

\section{Datasets and Metrics}
\label{sec:datasetsmetrics}
\noindent \textbf{Datasets.} 
We use five datasets.

\begin{itemize}
\item  \textbf{The Big Bang Theory (TBBT)} dataset~\cite{laught_machine} contains 228 episodes of  \textit{TBBT} TV show: (183,23,22) for (train,val,test). All episodes come with video, audio and subtitles, labelled as humor (or non) if followed (or not) by laughter.
\item \textbf{Multimodal Humor Dataset (MHD)}~\cite{humor_data} contains episodes from the TV show \textit{TBBT}, with 110 episodes split (84,6,20) for (train,val,test) (disjoint splits to TBBT). 
It contains multiple modalities; the subtitles are tagged as humor (or not).
\item \textbf{MUStARD}~\cite{mustard} contains 690 segments from four TV shows with video-audio-transcript labelled as sarcastic or not.
\item \textbf{UR-Funny}~\cite{urfunny} contains 1866 TED-talk segments with video-audio-transcript labelled as funny or not.
\item \textbf{Friends}~\cite{brown2021face,kalogeiton2020constrained} contains all 25 episodes ($\sim$23 minutes) from the third season of \textit{Friends} ($\sim$10 hours). We split them into 15 training  (1-15), 5 validation (16-20) and 5 test episodes (21-25).
Each episode comes with video, audio, face, body, voice tracks and features with speaker identifiers. 
In this work, we enrich this dataset by providing manually annotated laughter time codes. These annotated laughter time codes consist of time-stamps of the start and the end of all canned or not laughter. This results in 3.5k time-codes, with an average duration of 3 sec (0.3-16.5 sec), 138 average number of laughter per episode (109 to 182). 
The annotations are available: {\footnotesize{\url{https://www.lix.polytechnique.fr/vista/projects/2024_ijcv_liu/}}}.

\end{itemize}

\noindent \textbf{Metrics.} To evaluate \textbf{FunnyNet-W}, we use classification accuracy (Acc) and F1 score (F1). \\
For \textbf{laughter detector}, we use sample-scale at the detection level and frame-scale at the temporal level to compute precision (Pre), recall (Rec) and F1.

\section{Experiments}
\label{sec:exps}

In this section, we provide experiments for FunnyNet-W. 
First, we compare to the state of the art (Section~\ref{sub:sota}), then we provide an ablation of each component of FunnyNet-W (Section~\ref{sub:ablation}), and finally, we ablate our unsupervised laughter detector (Section~\ref{sub:expsaudio}).

\noindent \textbf{Implementation Details.} 
We train FunnyNet using Adam optimizer~\cite{adamw} with a learning rate of $1\times10^{-4}$, batch size of 32 and Pytorch~\cite{NEURIPS2019_9015}. The input audio is first downsampled by fixed sampling frequency (16000 Hz) and then transformed to log-scaled Mel spectrogram by mel-spaced frequency bins F = 64. 
At training, we use data augmentation: for frames, we randomly apply rotation and horizontal/vertical flipping, and randomly set the sampling rate to 8 frames; for audios, we apply random forward/backward time shifts and random Gaussian noises. For subtitles, we tokenize them as $max\_length=64$ inputs and send them to the language models.

\noindent \textbf{Setting.} In our experiments, we train FunnyNet-W on Friends. For MUStARD and UR-Funny, we fine-tune FunnyNet-W on their respective train sets. For TBBT and MHD, we fine-tune it only with a subset of the training set from TBBT (32 random episodes) These datasets come with data samples of uneven lengths. If the sample length is larger than 8 seconds (our best setting), we crop the last 8-second sequence to fit our model; otherwise, we pad zeros to the end. For UR-FUNNY,  we exclude from training the data samples with no sounds. \liu{For audio, textual, and visual encoders, we used the corresponding pre-trained models for feature extraction.} 

To understand the funny moments in the wild, we consider that subtitles do not come naturally with other modalities. Different from text-driven funny detections~\cite{mustard,hasan2021humor,urfunny}, we instead use off-the-shelf audio-to-text models, like WhisperX~\cite{bain2022whisperx}, to automatically generate texts from audios for funny detection. Hence, in addition to audiovisual data, we experiment with both real and automatically-generated texts (in the form of subtitles) for funny moment detection.


\subsection{Comparison to the State of the Art }
\label{sub:sota}

\begin{table*}[t]

\centering
\caption{
Comparison to the state of the art on five datasets. Modalities used per method A: audio, V: visual frames, $\text{T}^\text{{gt}}$: ground truth text (subtitles or transcript), $\text{T}^\text{{a}}$: automatically generated text (text extracted from speech), F: face.
\\
The column `Wild' signifies the methods that can run \emph{in the wild}, i.e., automatically without requiring ground truth information either for training or for testing. Note, most methods require ground truth labels (mostly in the form of textual subtitles or transcripts) both for training and testing. This is in contrast to FunnyNet-W which can automatically process videos in the wild. 
 \\
 $^{\dagger}$Reproduced results: we use the exact model as in~\cite{laught_machine}, pre-train it on Friends and fine-tune it on the other datasets 
}
		\begin{small}
		\resizebox{\linewidth}{!}{
\begin{tabular}{l c rr crr crr crr crr}
\toprule

\multirow{2}{*}{\textbf{Method / Metrics}} & \multirow{2}{*}{\textbf{Wild}} & \multicolumn{2}{c}{\cellcolor{mistyrose}{\textbf{TBBT}}} & \phantom{abc} &  \multicolumn{2}{c}{\cellcolor{mistyrose}{\textbf{MHD}}} & \phantom{abc} & \multicolumn{2}{c}{\cellcolor{mistyrose}{\textbf{MUStARD}}} & \phantom{abc} & \multicolumn{2}{c}{\cellcolor{mistyrose}{\textbf{UR-Funny}}} & \phantom{abc} & \multicolumn{2}{c}{\cellcolor{mistyrose}{\textbf{Friends}}} \\
&  & F1          & Acc     & \phantom{abc}   & F1         & Acc     & \phantom{abc}   & F1           & Acc & \phantom{abc}         & F1            & Acc & \phantom{abc}         & F1           & Acc          \\ 
\midrule
Random        &        --            & 46.3        & 50.0 & \phantom{abc}      & 56.1       & 50.9  & \phantom{abc}     & 48.3         & 48.7    & \phantom{abc}     & 50.2          & 50.2  & \phantom{abc}       & 51.0         & 51.0         \\
All positive     & --                & 60.3        & 43.2 & \phantom{abc}      & 75.6       & 60.8 & \phantom{abc}      & 66.7         & 50.0      & \phantom{abc}   & 75.4          & 50.7   & \phantom{abc}      & 66.7         & 50.0         \\
All negative        & --             & 0.0         & 56.8   & \phantom{abc}    & 0.0        & 39.2   & \phantom{abc}    & 0.0          & 50.0   & \phantom{abc}      & 0.0           & 49.3 & \phantom{abc}        & 0.0          & 50.0         \\
MUStARD 2019 (V+A+$\text{T}^\text{{gt}}$)~\cite{mustard}    & --              & -           & -  & \phantom{abc}        & -          & -  & \phantom{abc}        & 71.7         & 71.8    & \phantom{abc}     & -             & -    & \phantom{abc}        & -            & -            \\
MSAM 2021 (V+$\text{T}^\text{{gt}}$)~\cite{humor_data}       & --                & -           & -    & \phantom{abc}      & 81.3       & 72.4    & \phantom{abc}   & -            & -      & \phantom{abc}      & -             & -     & \phantom{abc}       & -            & -            \\
MISA 2020 (V+A+$\text{T}^\text{{gt}}$)~\cite{MISA}     & --                & -           & -    & \phantom{abc}      & -          & -     & \phantom{abc}     & -            & 66.2     & \phantom{abc}   & -             & 69.8     & \phantom{abc}    & -            & -            \\
HKT 2021 (V+A+$\text{T}^\text{{gt}}$)~\cite{hasan2021humor}        & --              & -           & -    & \phantom{abc}      & -          & -  & \phantom{abc}        & -            & 79.4    & \phantom{abc}     & -             & 77.4    & \phantom{abc}     & -            & -            \\
LaughM$^{\dagger}$ 2021 ($\text{T}^\text{{gt}}$)~\cite{laught_machine} & --  & 64.2        & 70.5 & \phantom{abc}      & \textbf{86.5}          & 76.3     & \phantom{abc}     & 68.6            & 68.7  & \phantom{abc}      & 71.9  & 67.6  & \phantom{abc} & 74.7 & 59.8     \\
FunnyNet (V+A+$\text{T}^\text{{gt}}$)   ~\cite{funnynet}     & --          & 73.8        & 75.8    & \phantom{abc}  & 83.4       & 78.6   & \phantom{abc}    & 79.5         & 79.9      & \phantom{abc}   & 84.1          & 79.9    & \phantom{abc}     & 88.2         & 85.8         \\
FunnyNet (V+F+A+$\text{T}^\text{{gt}}$)      ~\cite{funnynet}  & --           & 75.9        & 78.3 & \phantom{abc}      & 85.2       & 79.6 & \phantom{abc}      & 83.2         & 82.0     & \phantom{abc}    & 84.4          & 80.2  & \phantom{abc}       & 88.8         & 86.4         \\ 
\textbf{FunnyNet-W (V+A+$\text{T}^\text{{gt}}$)}     & --        &  \textbf{78.5}     &   \textbf{80.0}  & \phantom{abc}  & 84.6 & \textbf{80.1} & \phantom{abc}    & \textbf{85.9} &  \textbf{84.1}    & \phantom{abc}   & \textbf{84.5} & \textbf{80.2}  & \phantom{abc}     & \textbf{89.3}         & \textbf{86.7}        \\
\midrule
FunnyNet: V+F+A    ~\cite{funnynet}    & \checkmark        & 69.6        & 74.0 & \phantom{abc}     & \textbf{84.0}      & \textbf{79.3}  & \phantom{abc}     & \textbf{81.4} & \textbf{81.0}   & \phantom{abc}     & 83.7 & 78.0  & \phantom{abc}      & 86.8 & 84.8         \\
\textbf{FunnyNet-W: V+A+$\text{T}^\text{{a}}$}     & \checkmark            & \textbf{78.2}        & \textbf{79.1}    & \phantom{abc}  & 83.6       & 78.9   & \phantom{abc}    & 80.1         & \textbf{81.0}      & \phantom{abc}   & \textbf{84.2}          & \textbf{80.3}    & \phantom{abc}     & \textbf{88.2}         & \textbf{85.6}         \\

\bottomrule
\end{tabular}
			}
		\end{small}
	\label{tab:sota1}
\end{table*}

Here, we evaluate FunnyNet on five datasets: TBBT, MHD, MUStARD, UR-Funny and Friends and compare it to the state of the art: MUStARD~\cite{mustard}, MSAM~\cite{humor_data}, MISA~\cite{MISA}, HKT~\cite{hasan2021humor} and LaughM~\cite{laught_machine}. 
Table~\ref{tab:sota1} reports the results (including random, positive and negative baselines) for both metrics. We indicate the modalities each method uses as A: audio, V: video, $\text{T}^\text{{gt}}$: ground-truth text, $\text{T}^\text{{a}}$: automatically-generated text (speech-to-text), and F: face. Furthermore, we also indicate in the `Wild' column the methods that can run automatically without requiring ground-truth information (either for training or testing). Note that most methods require ground truth labels (mostly in the form of textual subtitles or transcripts) either for training or testing ($\text{T}^\text{{gt}}$). This is in contrast to FunnyNet-W, which can automatically process videos in the wild by exploiting speech-to-tex models ($\text{T}^\text{{a}}$).  

The first part of Table~\ref{tab:sota1} (no wild) demonstrates that overall the proposed  FunnyNet-W (V+A+$\text{T}^\text{{gt}}$) outperforms all methods 
on all five datasets. 
For TBBT it outperforms the LaughM by a notable margin of +10\% for F1 and Acc, and FunnyNet by +3\% in both metrics. For MHD, it outperforms MSAM  by 3\% in F1 and 7\% in Acc, LaughM and FunnyNet by 3\% and 1\%, respectively, in Acc. Furthermore, FunnyNet-W outperforms MUStARD, MISA, HKT, LaughM and FunnyNet by \liu{3-12\%} in F1 and 2-15\% in Acc for MUStARD and 1-15\% in F1 and 1-10\% in Acc for UF-Funny. 
For Friends, we observe similar patterns, where we outperform LaughM by \liu{15\%} in F1 and \liu{26\%} in Acc and FunnyNet by approximately +1\% in both F1 and Acc. 
These results confirm the effectiveness of FunnyNet compared to other methods. 

The major advantage and motivation of FunnyNet-W and its predecessor FunnyNet~\cite{funnynet} is the fact that they can run \emph{in the wild}, i.e. without requiring ground truth data either at training or test time. To this end, the second part of Table~\ref{tab:sota1} reports results when experimenting in the wild. We observe that for TBBT, remarkably FunnyNet-W outperforms its predecessor FunnyNet by 5-10\% in F1 and Acc, while for MHD it is inferior by 1\% \liu{or similar for MUStARD}.
For UR-Funny and Friends, FunnyNet-W outperforms FunnyNet consistently by 1-3\% in all metrics. When we compare FunnyNet-W-$\text{T}^\text{{a}}$ to the first part of the table, we observe that it still produces on par or superior results to all other methods. This clearly shows the superiority of FunnyNet-W even when compared to methods that have access to manually annotated ground-truth data. 

Our remarks are: 
First, FunnyNet-W outperforms most methods in both metrics in both settings, when using ground truth text ($\text{T}^\text{{gt}}$) or when being \emph{in the wild} ($\text{T}^\text{{a}}$). Second, the performance in the out-of-domain UR-Funny is significantly high.  
Third, for TBBT and MHD our results are much less optimized than the ones from LaughM or MSAM, as we do not have access to the exact same test videos as either work, so inevitably there are some time shifts or wrong labels\footnote{The label time shift is 0.3-1s on TBBT and 0.3-2s on v2.} and we use much fewer training data (32 vs 183 episodes in LaughM vs 84 episodes in MHD). These highlight that FunnyNet-W is an effective model for funny moment detection.

Note that in the remainder of this work, unless stated otherwise, using the automatically-generated text (in the form of subtitles) is the \textbf{default} setting of FunnyNet-W. For simplicity, we denote the $\text{T}^\text{{a}}$ by T.

\subsection{Ablation of FunnyNet-W}
\label{sub:ablation}
In this section, we provide ablations of FunnyNet-W. 
Specifically we ablate the encoders (Section~\ref{subsub:encoders}), the modalities (Section~\ref{subsub:modalities}), the cross attention fusion module (Section~\ref{subsub:caf}), the length of input videos (Section~\ref{subsub:time}) and the losses (Section~\ref{subsub:losses}). 

\begin{table*}[t]
\begin{minipage}[b]{0.32\textwidth}

\centering
\caption{Ablation of visual encoders on Friends. 
}
\resizebox{0.95\linewidth}{!}{
    \begin{tabular}{ccc|cc}
    \toprule
    \multicolumn{3}{c|}{\textbf{Modality}} & \multirow{2}{*}{F1} & \multirow{2}{*}{Acc} \\ 
    \cellcolor{mistyrose}{A} & \cellcolor{mistyrose}{V} & \cellcolor{mistyrose}{$\text{T}^\text{{a}}$}   &  &  \\
    \midrule
    \multirow{2}{*}{BYOL-A} & Timesformer & \multirow{2}{*}{Bert} & 84.2 & 80.9 \\
    & VideoMAE &  & 85.3 & 82.3 \\
    \toprule
    \cellcolor{mistyrose}{A} & \cellcolor{mistyrose}{V} & \cellcolor{mistyrose}{$\text{T}^\text{{gt}}$} &  &  \\
    \midrule
    \multirow{2}{*}{BYOL-A} & Timesformer & \multirow{2}{*}{Bert} & 84.9 & 80.8 \\
    & VideoMAE &  & 87.2 & 83.8 \\

    \bottomrule
    
    \end{tabular}}
    \label{tab:ablation_V_encoder}
\end{minipage}
\hfill
\begin{minipage}[b]{0.3\textwidth}
\centering
\caption{Ablation of text encoders on Friends.
}
    \resizebox{0.95\linewidth}{!}{
    \begin{tabular}{ccc|cc}
    \toprule
    \multicolumn{3}{c|}{\textbf{Modality}} & \multirow{2}{*}{F1} & \multirow{2}{*}{Acc} \\ 
    \cellcolor{mistyrose}{A} & \cellcolor{mistyrose}{V} & \cellcolor{mistyrose}{$\text{T}^\text{{a}}$} &  &  \\
    \midrule

     \multirow{3}{*}{BYOL-A} & \multirow{3}{*}{VideoMAE} & Bert & 85.3 & 82.3 \\
     &  & GPT2 & 85.2 & 82.3 \\
     &  & LlaMa-2 & 88.2 & 85.6 \\
     
    \bottomrule
    
    \cellcolor{mistyrose}{A} & \cellcolor{mistyrose}{V} & \cellcolor{mistyrose}{$\text{T}^\text{{gt}}$} & F1 & Acc  \\
    \midrule

    \multirow{3}{*}{BYOL-A} & \multirow{3}{*}{VideoMAE} & Bert & 87.2 & 83.8 \\
     &  & GPT2 & 88.1 & 85.6 \\
     &  & LlaMa-2 & 89.3 & 86.8 \\

    \bottomrule
    \end{tabular}}
	\label{tab:ablation_T_encoder}
\end{minipage}
\hfill
\begin{minipage}[b]{0.32\textwidth}
\centering
\caption{Ablation of audio encoders on Friends. 
}
    \resizebox{0.95\linewidth}{!}{
    \begin{tabular}{ccc|cc}
    \toprule
    \multicolumn{3}{c|}{\textbf{Modality}} & \multirow{2}{*}{F1} & \multirow{2}{*}{Acc} \\ 
    \cellcolor{mistyrose}{A} & \cellcolor{mistyrose}{V} & \cellcolor{mistyrose}{$\text{T}^\text{{a}}$} &  &  \\
    \midrule
    BEATS & \multirow{4}{*}{VideoMAE} & \multirow{4}{*}{LlaMa-2} & 78.2 & 65.1 \\
    CAV-MAE &  &  & 87.3 & 83.8 \\
    BYOL-A-v2 &  &  & 87.6 & 84.7 \\
    BYOL-A &  &  & 88.2 & 85.6\\
    \bottomrule
    \end{tabular}}

	\label{tab:ablation_A_encoder}
 
\end{minipage}

\caption*{
\small{A: audio, V: visual frames, $\text{T}^\text{{gt}}$: ground truth text, $\text{T}^\text{{a}}$: automatically generated text (text extracted from speech)}}

\end{table*}

\begin{table*}[t]
\begin{minipage}[b]{0.35\textwidth}
\centering
	\caption{
	Ablation of modalities of FunnyNet-W on Friends test set.
	}
    \begin{tabular}{lll|rr}
    \toprule
    \multicolumn{3}{c|}{\textbf{Modality}} & \multirow{2}{*}{F1} & \multirow{2}{*}{Acc} \\
    \cellcolor{mistyrose}{V} & \cellcolor{mistyrose}{A} & \cellcolor{mistyrose}{T} & & \\
    \midrule
     \checkmark & -- & --  & 73.2 & 64.1 \\
    -- & \checkmark & -- & 73.7 & 66.6 \\
    -- & -- & \checkmark & 77.8 & 68.1 \\
    \checkmark & \checkmark & -- & 84.3 & 79.3 \\
    -- & \checkmark & \checkmark & 84.5 & 80.3 \\
    \checkmark & -- & \checkmark & 74.9 & 64.3 \\
    \checkmark & \checkmark & \checkmark & \textbf{88.2} & \textbf{85.6} \\
     \bottomrule
    \end{tabular}
	\label{tab:ablation}

\end{minipage}
\hfill
\begin{minipage}[b]{0.6\textwidth}

\centering
\caption{
   Ablation of CAF of FunnyNet-W on Friends test set. (A: audio, V: visual frames, T: text) }
    \begin{tabular}{cc r@{\hspace{3pt}}r r@{\hspace{3pt}}r r@{\hspace{3pt}}r r@{\hspace{3pt}}r}
    \toprule
    \multicolumn{2}{c}{\cellcolor{mistyrose}{\textbf{CAF}}}   & \multicolumn{2}{c}{\cellcolor{mistyrose}{\textbf{A+V}}} & \multicolumn{2}{c}{\cellcolor{mistyrose}{\textbf{A+T}}} & \multicolumn{2}{c}{\cellcolor{mistyrose}{\textbf{V+T}}} & \multicolumn{2}{c}{\cellcolor{mistyrose}{\textbf{A+V+T}}} \\
    \textbf{Self} & \textbf{Cross} & F1 & Acc    & F1 & Acc & F1 & Acc & F1 & Acc \\ 
    \midrule
    - & -     & 80.1           & 76.5           & 81.0          & 76.9          & 73.5             & 63.8             & 82.4             & 77.8             \\
    \checkmark & -      & 81.1           & 77.3           & 81.4          & 77.5          & 74.4             & 64.4             & 85.7             & 81.8             \\
    - & \checkmark     & 83.6           & 78.7           & 82.3          & 78.7          & 74.6             & 64.2             & 85.4             & 81.4             \\
    \checkmark & \checkmark      & \textbf{84.3}           & \textbf{79.3}           & \textbf{84.5}           & \textbf{80.3}          & \textbf{74.9}             & \textbf{64.3}             & \textbf{88.2}             & \textbf{85.6}             \\ 
    \midrule
    \multicolumn{2}{c}{MMCA~\cite{cross_attn_2}}      &   83.1         &   78.3         &      83.4     &    79.8       &  73.6     &   63.8           &  87.0     &   84.5           \\
    \multicolumn{2}{c}{CoMMA~\cite{tanCOMMA2021}}      &     83.5       &  78.5    &   83.9    &    \textbf{80.3}   &   74.2    &  64.1           &  87.6    &  85.1           \\
    \bottomrule
    \end{tabular}
	\label{tab:ablation_caf}
	
\end{minipage}
\end{table*}

\subsubsection{Ablation of Encoders}
\label{subsub:encoders}

\paragraph{Visual encoder.}
Table~\ref{tab:ablation_V_encoder} ablates two video encoders on Friends, i.e. Timesformer~\cite{timesformer} and VideoMAE~\cite{tong2022videomae} for two scenarios: one using automatically generated text ($\text{T}^\text{{a}}$) and when using ground-truth text ($\text{T}^\text{{gt}}$). Given the same video sequence, we use the best settings for them (8 frames for Timesformer and 16 frames for VideoMAE). We observe that using VideoMAE outperforms Timesformer by about 1-3\% in F1 score and 2-3\% in Acc. This is expected because VideoMAE is a larger model, and it also uses a masked autoencoder for unsupervised learning; hence it can generalize better than Timesformer. When comparing the results between using ground truth and automatically generated texts, we observe that the improvements of using VideoMAE are consistent, and the differences are very small (1-2\% in both F1 and Acc).

\paragraph{Text encoder.} 
Table~\ref{tab:ablation_T_encoder} ablates three different text encoders in FunnyNet-W on Friends: Bert~\cite{bert}, GPT2~\cite{gpt2} and LlaMa-2~\cite{llama2} (7B model) for two scenarios (one using automatically generated text ($\text{T}^\text{{a}}$) and when using ground-truth text ($\text{T}^\text{{gt}}$)). Given the other ablation study, we choose VideoMAE and BYOL-A as the best visual and audio encoders, respectively. We observe that using LlaMa-2 gives the best improvements in both F1 and Acc. Interestingly, using GPT2 results in inferior performance than using Bert. This finding is consistent with what we observe in LLM models. For LlaMa-2, we note that the differences between using ground truth and automatically generated texts are minor, about 0.8-1\% in both F1 and accuracy.

\paragraph{Audio encoder.}
Table~\ref{tab:ablation_A_encoder} ablates four audio encoders on Friends: Beats~\cite{chen2022beats}, CAV-MAE~\cite{gong2023cavmae}, BYOL-A-v2~\cite{byola_v2} and BYOL-A~\cite{byol_a}. 
Given the previous ablation studies, we choose VideoMAE and LlaMa-2 as the best visual and text encoders and operate directly with automatically generated text ($\text{T}^\text{{a}}$). The results show that CAV-MAE, BYOL-A-v2 and BYOL-A perform on par (approximately 1\% difference in F1 and Accuracy). In our experiments, we use BYOL-A as it results in the best F1 and Accuracy but it also requires fewer parameters than the other models. 

\paragraph{Subtitles sources.}
Tables~\ref{tab:ablation_V_encoder} and \ref{tab:ablation_T_encoder} report results for two scenarios: one using automatically generated text ($\text{T}^\text{{a}}$) and when using ground-truth text ($\text{T}^\text{{gt}}$). Consistently, we observe that using the ground truth text outperforms using the automatically-generated one. This is expected, as $\text{T}^\text{{a}}$ includes imperfect transcripts. We note, however, that the difference in both F1 and Accuracy are minor (1-3\% for both metrics). This highlights that substituting ground-truth with an automatic speech-to-text model is a good trade-off between good performance and the ability to run in the wild, i.e., without requiring manual ground truth labels. 

\subsubsection{Ablation of Modalities}
\label{subsub:modalities}

Table~\ref{tab:ablation} ablates all modalities of FunnyNet-W on the Friends test set. Using text alone (\liu{third} row) produces better results than when using the visual or audio modality alone (first and second rows). This highlights the efficiency of large dataset pre-training and the representation power of Large Language Models (since we use LlaMa 2 as the textual encoder). Using audio alone (second row) leads to the second-best performance compared to using single modalities, underlying that audio is more suitable than visual cues for our task, as it encompasses the way of speaking (tone, pauses). 
Combining modalities outperforms using single ones: combining visual and audio (fourth row) or visual and text (sixth row) increases the F1 by approximately \liu{1.3--10\%} and the Acc by \liu{0.2--15\%}. 
This is expected as audio or text bring complementary information to the visual modality~\cite{morgado2021audio,clip} and their combination helps discriminate funny moments.  
Combining audio and text (fifth row) leads to \liu{larger boosts than audio+visual or text+visual (fourth and sixth rows), as audio and text contain complementary information regarding character dialogues, expression in voices and background music. }
Overall, using all modalities achieves the best performance.

\subsubsection{Ablation of Cross-Attention Fusion (CAF)} 
\label{subsub:caf}

Table~\ref{tab:ablation_caf} reports results with various cross- and self-attention fusions in CAF. 
We observe that including either self- or cross-attention (second, third rows) brings improvements over not having any (first row), indicating that they enhance the feature representation. 
The fourth row shows that using them both for feature fusion leads to the best performance. 
For completeness, we also compare CAF against the \liu{state of the art:} MMCA~\cite{cross_attn_2} and CoMMA~\cite{tanCOMMA2021}. All CAF, MMCA and CoMMA use self and cross-attentions jointly for feature extraction. Their main difference is that both MMCA and CoMMA first use self-attention to individually process each modality, then concatenate all modalities together and process them using cross-attention to output the final feature representation. Instead, CAF uses cross-attention to gradually fuse one modality with the rest of the modalities to fully explore cross-modal correlations. 
The results (fourth, fifth, and last rows) show that CAF outperforms MMCA~\cite{cross_attn_2} and CoMMA~\cite{tanCOMMA2021} by 0.1--0.4 in F1 score and 0.03--0.2 in accuracy. This reveals the importance of the gradual modality fusion, and hence the superiority of CAF.

\begin{table}
\caption*{\textbf{Varying time window lengths}
}
\begin{minipage}[b]{\linewidth}
\centerline{\includegraphics[width=\columnwidth]{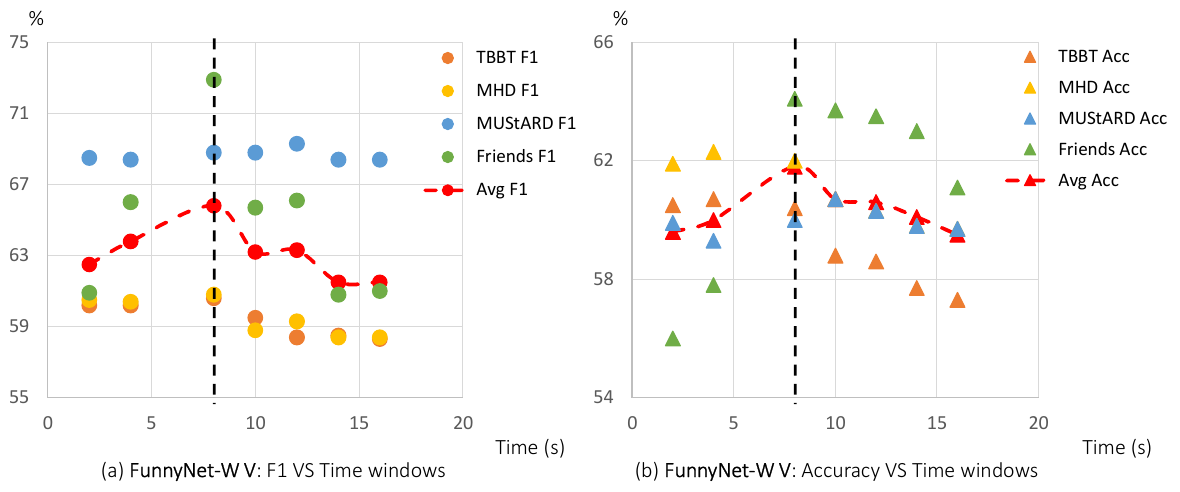}} 
\end{minipage}
\hspace{0.15cm}
\par
\centering
\centerline{\includegraphics[width=\columnwidth]{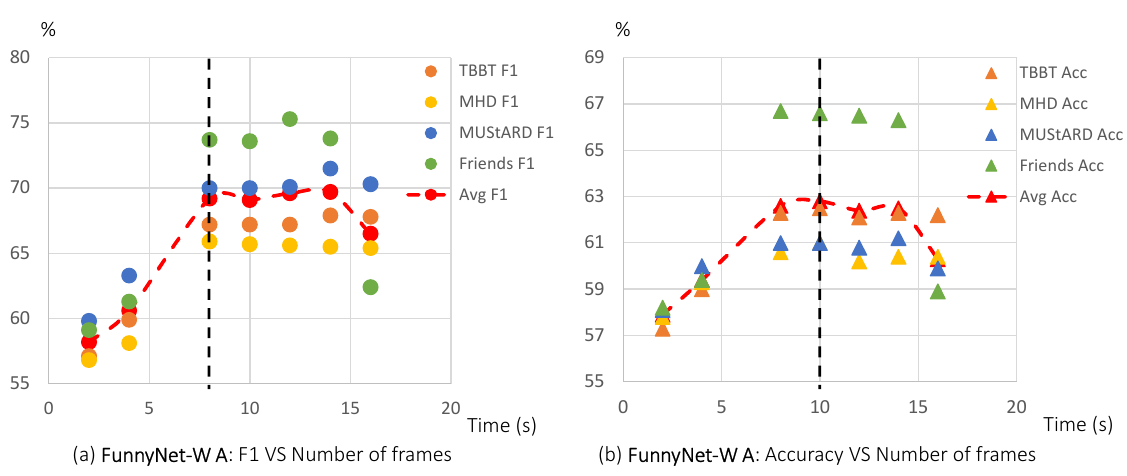}} 

\captionof{figure}{Comparison of various time window lengths used as input of the (top) visual encoder of FunnyNet-W (referred to as FunnyNet-W V) and (bottom) audio encoder of FunnyNet-W (referred to as FunnyNet-W A). We illustrate \textit{(left,a)} the F1 score and \textit{(right,b)} the accuracy on different datasets. The average results are plotted in red lines.}
\label{fig:v_a_time_window}
\end{table}

\subsubsection{Impact of time}
\label{subsub:time}

In this section, we examine the impact the length of the time window has on the final results, as well as the number of sampled frames within the time window. 

\paragraph{Influence of Time Window.} 
Following~\cite{timesformer}, our proposed FunnyNet-W is trained on fixed-length inputs of multiple modalities that last 8 seconds. Here, we examine the impact that the length of time window has on FunnyNet-W and illustrate results on four datasets (as well as their average in a dashed red line) in 
Figure~\ref{fig:v_a_time_window}. 
For this, we use input time windows of varying lengths (from 2 to 16 seconds) in either the visual encoder of FunnyNet-W (referred to as FunnyNet-W V, top in Figure~\ref{fig:v_a_time_window}) or the audio encoder of FunnyNet-W (referred to as FunnyNet-W A, bottom in Figure~\ref{fig:v_a_time_window}). 

When ablating the input length of the visual input (top in Figure~\ref{fig:v_a_time_window}), we observe that using approximately 8 seconds achieves the best performance compared to all other settings. 
Specifically, for F1 (a, left), we observe that for all datasets, the best result is achieved when using 8 seconds, whereas the second and third results are achieved when using 10 and 12 seconds length of the input. 
For Accuracy (b, right), the performance follows the same trend: the best accuracy is reached for 8-second inputs, while the 10 and 12-second inputs reach the second and third-best accuracies.
Interestingly, for both F1 and Accuracy, for the average amongst all datasets (red dashed lines), we observe that both metrics degrade when using longer visual input windows (e.g. more than 15 seconds). 
This is probably because longer inputs contain too much visual or audio information across both positive and negative samples, which confuses the model and leads to more incorrect predictions.

When ablating the input length of the audio input (bottom in Figure~\ref{fig:v_a_time_window}), we observe that similar to the previous conclusions, the time window of 8 seconds leads to the best performance both in F1 and Accuracy. Nevertheless, using a longer time window improves the prediction accuracy, in contrast to the visual ablation. Specifically, the best time window setting is between 8 and 12 seconds. For any time windows outside this range, the performance is getting worse.

In our experiments, we use a time window of 8 seconds as a good trade-off between the performance of the visual and audio encoders. 

\begin{figure}[t]
\centering
    \centerline{\includegraphics[width=\columnwidth]{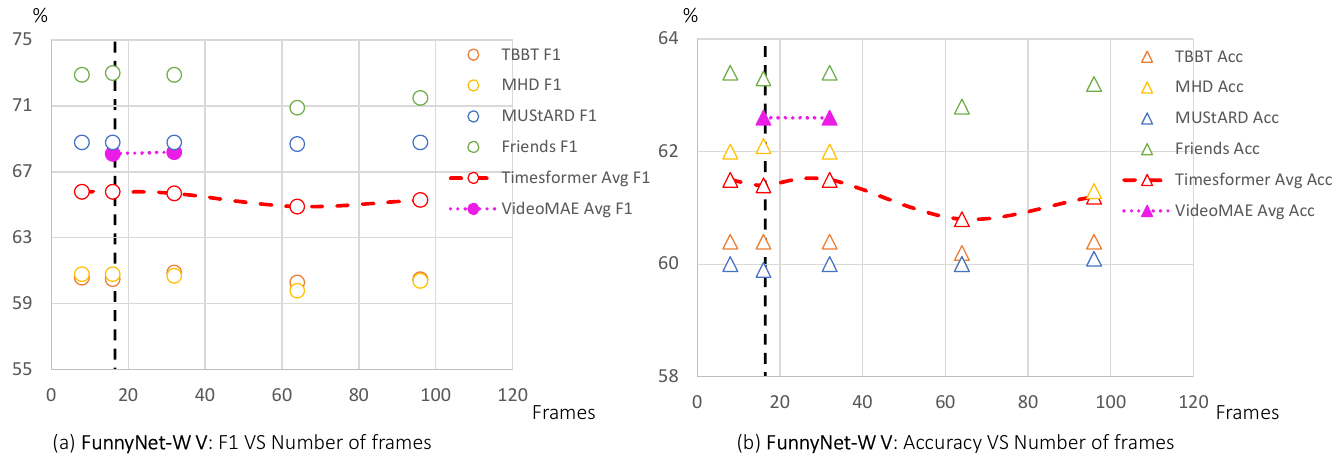}} 
		\caption{
    	Comparison of different lengths of time windows for the visual encoder of FunnyNet-W (referred to as FunnyNet-W V). We illustrate \textit{(a)} the F1 score and \textit{(b)} the accuracy on different datasets. The average results are plotted in red points and lines for Timesformer and magenta for VideoMAE. 
		 }
		\label{fig:v_num_frame}
\end{figure}

\paragraph{Influence of Sampled Frames.} 
Given the input time window of 8 seconds, we test the scenario where we sample different numbers of frames within a fixed 8-second time window. In particular, we examine the impact when sampling from 8 to 100 frames. The results are shown in Figure~\ref{fig:v_num_frame}, where we illustrate the (left, a) F1 score and (right, b) accuracy over the number of sampled frames. Our results suggest that the number of frames has no or only a trivial impact on the final performance. This is expected, since sampling more frames in a fixed time window mainly produces redundancy without introducing new relevant information. Furthermore, in this ablation, we also compare the results obtained when using Timesformer (red points and dashed line) and VideoMAE (magenta points and dashed line) in the visual encoder. We observe that using VideoMAE outperforms Timesformer in all settings, hence the final FunnyNet-W uses VideoMAE for the visual encoder.

\begin{table}[t]
    \caption{Ablation of losses used to train FunnyNet-W.
	}
    \centering
        \begin{small}
\begin{tabular}{cccc|rr}
\toprule

$L_{\text{cls}}$ & $L_{\text{ss}}$ & $L_{\text{koleo}}$ &  $L_{\text{clip}}$ & F1 & Acc \\
\midrule
\checkmark & -- & -- & -- & 70.9 & 68.0 \\
\checkmark & \checkmark & -- & --  & \textbf{88.2} & \textbf{85.6} \\ 
\checkmark & \checkmark & \checkmark & --  & 86.7 & 84.6 \\ 
\checkmark & \checkmark & -- &\checkmark  & 87.3 & 84.6 \\
\checkmark & \checkmark & \checkmark   & \checkmark  
 & 85.4 & 82.2 \\
 \bottomrule
\end{tabular}
\end{small}
\label{tab:loss}
\end{table}

\subsubsection{Ablation of losses}
\label{subsub:losses}

FunnyNet-W uses the classification $L_{\text{cls}}$ and the self-supervised contrastive losses $L_{\text{ss}}$. 
Here, we examine their impact by training FunnyNet-W with and without $L_{\text{ss}}$. 
Table~\ref{tab:loss} reports the results on Friends, where we observe that adding $L_{\text{ss}}$ improves over +10 in all metrics (first two rows). 
This reveals that using the auxiliary self-supervised task of syncing audiovisual data helps to identify the funny moments in videos. 

Recently, Koleo~\cite{sablayrolles2019spreading}($L_{\text{koleo}}$)and CLIP~\cite{clip} ($L_{\text{clip}}$) have been proposed for improving unsupervised feature clustering. To examine the impact of these two losses, we train FunnyNet-W with different loss combinations and show the results in Table~\ref{tab:loss}. 
We observe that including Koleo and/or CLIP losses (third-fifth rows) results in a small drop in both F1 and accuracy compared to the proposed loss configuration (second row). Regarding the Koleo loss, this drop is probably because Koleo encourages a uniform span of the features within a batch which maximizes the variances of features and affects the binary decisions on the boundaries. Regarding the CLIP loss, the drop can be explained by the fact that CLIP is widely used for multi-class feature projection, which may complicate the funny or not-funny classification

\begin{table}[t]
\centering
\caption{Comparison to the state of the art of FLOPs count (FLOPs), number of parameters (Params) and inference runtime average (Runtime). We report two versions of model complexity, FunnyNet (V+F+A) and FunnyNet-W (V+F+T) with including pre-trained encoders, and FunnyNet$^b$ and FunnyNet-W$^b$ without including pre-trained encoders. 
}
\begin{small}
    \resizebox{\linewidth}{!}{
    \begin{tabular}{l|rrr}
    \toprule
    \multirow{ 2}{*}{\textbf{Model}} & \textbf{FLOPs} & \textbf{Params} & \textbf{Runtime} \\ 
     & \textbf{($10^9$)} & \textbf{ ($10^6$)} & (ms) \\ 
    \midrule
    MISA 2020 (V+A+T)~\cite{MISA} & 138.8 & 111.2 & 33.64 \\
    HKT 2021 (V+A+T)~\cite{hasan2021humor} & 7.6 & 16.8 & 25.91 \\
    LaughM 2021 (T)~\cite{laught_machine} & 2.5 & 112.4 & 11.15 \\
    \midrule
   FunnyNet$^b$ & 4.4 & 39.5 & 45.25 \\
    FunnyNet (V+F+A) & 190.9 & 126.9 & 45.25 \\
    FunnyNet-W$^b$ & 3.1 & 29.6 & 30.21 \\
    FunnyNet-W (V+F+T) & 72190 & 70228 & 30.21 \\
    \bottomrule
    \end{tabular}}
\end{small}
	\label{tab:profile_models}
\end{table}
\paragraph{Model complexity.}
We also compare in Table~\ref{tab:profile_models} the complexity of FunnyNet and FunnyNet-W to the other state-of-the-art models. Note that both models use pre-trained visual, audio and text encoders. For completion, we also report the metrics when including the complexity of the visual, audio and text backbone encoders. We observe that the gain in performances and the unsupervised aspect of FunnyNet-W impacts its complexity. Indeed, FunnyNet-W is a huge model, with an increase of approximately $52$ GFLOPS, $16$M of parameters and $11$ms on runtime, in comparison to the second-heaviest model~\cite{MISA}. Additionally, when comparing FunnyNet to FunnyNet-W, the latter replaces the face encoder with a text encoder and uses larger visual and text encoders, VideoMAE and LlaMa2, respectively. These lead to higher complexity on GFLOPS and parameters. However, the overall inference time is reduced because it does not require online per-frame face detection, masking, and feature extraction.

\begin{table*}[t]
    \centering
    \caption{
    Evaluation of Laughter Detection on Friends.
    We compare five versions of our laughter detector, denoted as `Ours', employing different feature encoders, along with two external audio laughter detectors. 
    The last row corresponds to the actual configuration used in FunnyNet-W.
    }
    \begin{tabular}{l r@{\hspace{7pt}}r@{\hspace{7pt}}r@{\hspace{7pt}}r r r@{\hspace{7pt}}r@{\hspace{7pt}}r r r@{\hspace{7pt}}r@{\hspace{7pt}}r}
    \toprule
    & \multicolumn{4}{c}{\cellcolor{mistyrose}{\textbf{Temporal}}} &  & \multicolumn{3}{c}{\cellcolor{mistyrose}{\textbf{Det} IoU = 0.3}} & & \multicolumn{3}{c}{\cellcolor{mistyrose}{\textbf{Det} IoU = 0.7}} \\
    & Acc & Prec & Rec & F1 & \phantom{abc} & Prec & Rec & F1 & \phantom{abc} & Prec & Rec & F1 \\
    \midrule

    LD~\cite{old_laughter_detector} & 43.6 & 35.7 & \textbf{99.0} & 52.3 & & 25.7 & 22.1 & 23.4 & & 4.0 & 3.7 &	3.8 \\
    
    RLD~\cite{new_laughter_detector} & 74.5 & 58.9 & 62.0 & 59.7 &  & 66.2 & 53.7 & 59.1 & & 18.5 & 15.0 &	16.5 \\
    
    \midrule
    
    Ours Wav2CLIP\cite{wav2clip} & 77.6 & 64.5 & 63.7 & 63.7 & & 91.3 & 61.2 & 73.1 & & 49.7 & 33.5 & 39.9 \\
    
    Ours CAV-MAE\cite{gong2023cavmae}  & 84.3 & 75.4 & 74.2 & 74.6 & & 92.3 & 80.0 & 85.6 & & 52.2 & 45.5 & 48.4 \\
    
    Ours BYOL-A\cite{byol_a} & 86.0 & 76.9 & 79.4 & 77.8 & & 94.6 & 82.3 & \textbf{87.8} & & 54.1 & 47.1 &	50.3 \\
        
    Ours BYOL-A-v2\cite{byola_v2} & 82.1 & 67.7 & \textbf{82.6} & 74.1 & & 92.5 & \textbf{83.5} & 87.6 & & 51.9 & 47.0 & 49.2 \\
    
   Ours BEATs\cite{chen2022beats} & \textbf{86.4} &\textbf{78.4} &{78.3} &\textbf{78.1} &{} &\textbf{95.2} &{81.6} &{87.7} &{} &\textbf{55.1} &\textbf{47.3} &\textbf{50.8} \\
    
    \bottomrule
    \end{tabular}
    \label{tab:audio_results}
\end{table*}

\subsection{Analysis of Unsupervised Laughter Detector}
\label{sub:expsaudio}

\paragraph{Comparison to the state of the art.}
We compare our laughter detector with the \liu{state of the art:} LD~\cite{old_laughter_detector} laughter detector used in \cite{mustard} and RLD~\cite{new_laughter_detector}. 
The results on the Friends dataset are presented in Table \ref{tab:audio_results}. 
Overall, our detector demonstrates superior performance compared to both supervised methods. Notably, our detector combined with BEATs features consistently demonstrates superior performance, excelling for instance in temporal precision ($78.4\%$), and detection precision for both thresholds ($95.2\%$ for $0.3$ and \liu{$55.1\%$} for $0.7$). Our method combined with BYOL-A and BYOL-A-v2 features also showcases a balanced performance, maintaining high temporal accuracy ($86.0\%$ and $82.1\%$ respectively). In comparison, LD exhibits high temporal recall ($99.0\%$) but lower temporal precision ($35.7\%$) highlighting a bias in its predictions. While RLD achieves a better balance between temporal precision and recall ($58.9\%$ $62.0\%$ respectively) it is still far from our results.

Furthermore, we evaluate our detector using five audio feature extractors: Wav2CLIP~\cite{wav2clip}, CAV-MAE~\cite{gong2023cavmae}, two versions of BYOL-A~\cite{byol_a,byola_v2}, and BEATs~\cite{chen2022beats}. Among these, the BEATs encoder exhibits the most suitable audio representation capacity for our detector, providing the best results (last row).
During the analysis of the laughter detection, we make three important observations:
(i) The majority of false positives are unfiltered sounds that are not easily separable using K-means clustering.
(ii) The majority of false negatives correspond to intra-diegetic laughter, which is typically less loud and therefore more challenging to detect.
(iii) The peak detector fails in scenarios where music overlaps with laughter, such as in party settings.

\begin{figure}[t]
\centering
		\centerline{\includegraphics[width=\columnwidth]{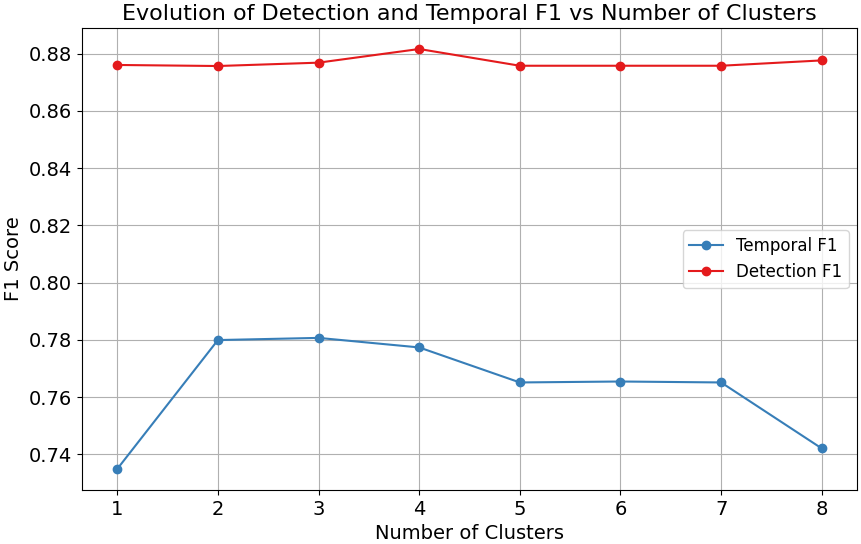}}
		\caption{\small{
		Evolution of the temporal (\textit{blue}) and detection (\textit{red}) F1 scores according to the number of clusters chosen for the K-means algorithm at the end of the laughter detection pipeline}}
		\label{fig:cluster_f1}
\end{figure}

\begin{figure*}[t]
\centering
		\centerline{\includegraphics[width=\textwidth]{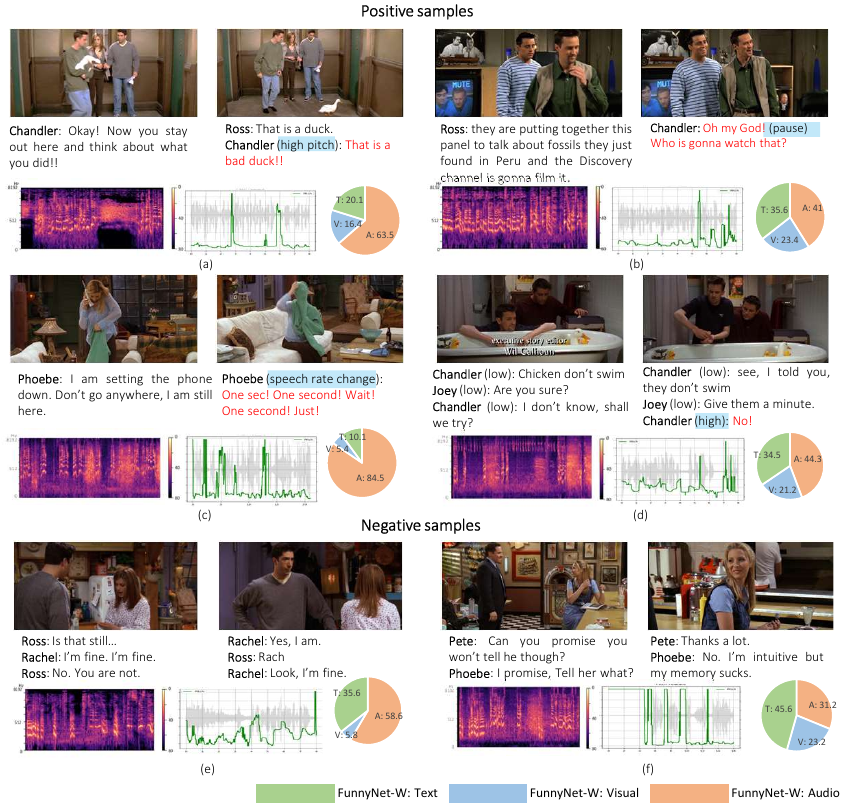}} 
		\caption{
		\small{
        Visualization of (\textit{a},\textit{b},\textit{c},\textit{d}) funny, (\textit{e},\textit{f}) non-funny predictions on the Friends test set.
		We show the audio, visual and text inputs, the learned average weights of cross-attentions from CAF (\textit{pie chart}), and the subtitles (for better understanding). 
		}}
		\label{fig:vis_cls}
\end{figure*}

\paragraph{Influence of Clustering on the Detection Performance.} 
Here, we examine how the choice of the cluster count parameter $K$ in the K-means algorithm influences the performance of our laughter detector. 
In practice, laughter chunks significantly outnumber music chunks. Consequently, in the third stage of Figure~\ref{fig:audio_pipeline}, we exclude the smallest cluster— identified as the music cluster through empirical assessment— and retain the clusters comprising the laughter chunks.

Figure~\ref{fig:cluster_f1} shows the performance of the detection pipeline both at the detection level (red lines) and at the temporal level (blue lines) as a function of different numbers of clusters (x-axis). 
Overall, we make the following three observations: (1) For 1 cluster, we note that using one cluster is equivalent to no clustering. (2) Between 2 and 4 clusters, we note that F1 scores are higher than for 1 cluster. Here, there are enough degrees of freedom for the K-means algorithm to correctly detect the centroid of the music cluster. (3)For more than 7 clusters, we note that F1 scores tend to converge to the same value as for 1 cluster. Here, there are too many degrees of freedom for the K-means algorithm, and therefore it detects multiple centroids for the music cluster. Thus, the higher the number of clusters, the smaller the music sub-cluster we have, with the extreme case of having one cluster per sample, thus having the same effect as no clustering.

Moreover, Figure~\ref{fig:cluster_f1} shows that the detection F1 score (red line) is less sensitive to the number of clusters than the temporal F1 score (blue line). This can be explained by the fact that music chunks are generally longer than laughter chunks. Thus, by removing longer false positive chunks, we improve temporal metrics, whereas the impact is less important at the sample scale for detection metrics.

\section{Analysis of FunnyNet-W}
\label{sec:analysis}

\begin{figure}[t]
\centering
		\centerline{\includegraphics[width=\columnwidth]{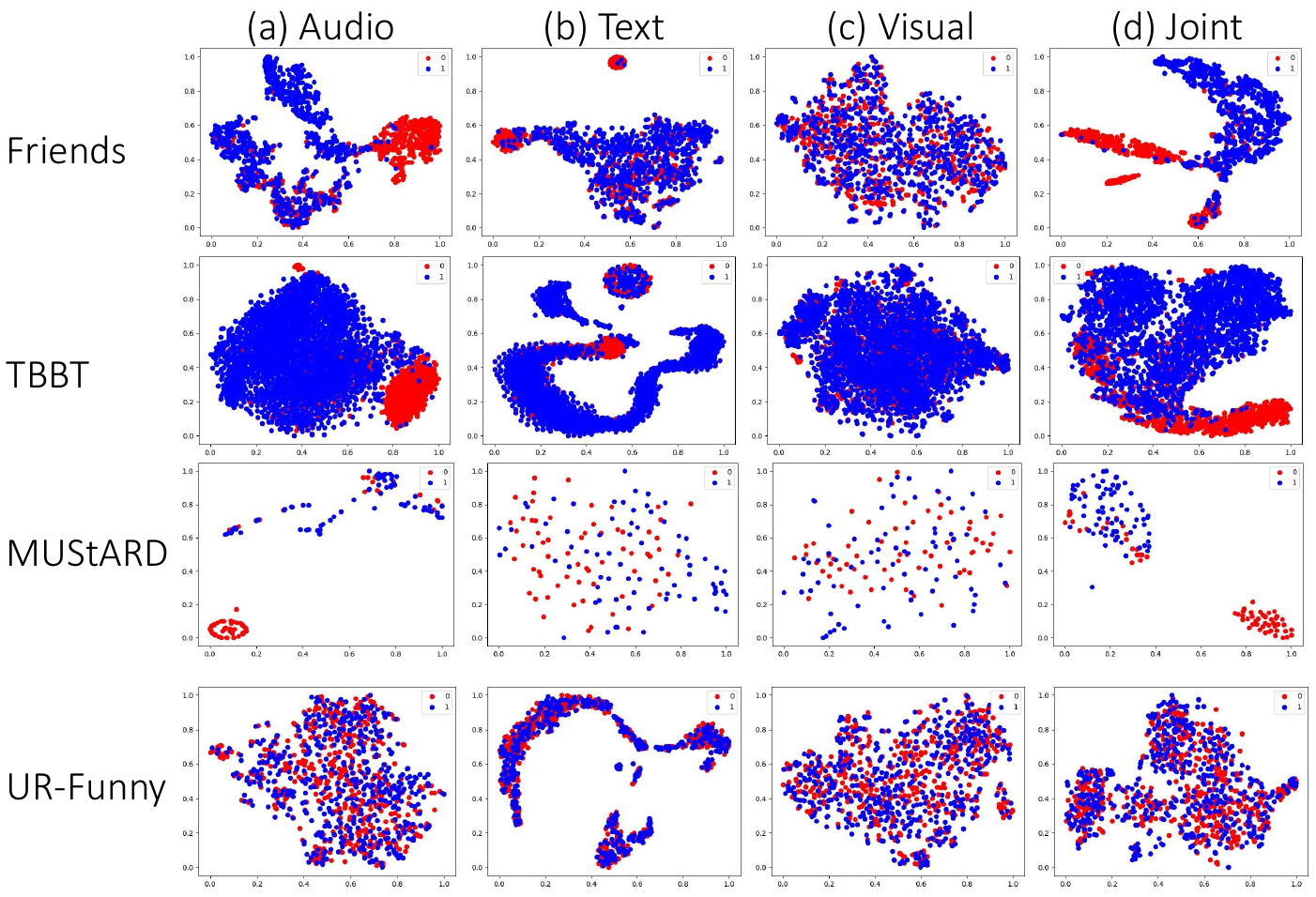}} 
		\caption{
		t-SNE visualization of embeddings on Friends for (\textit{a}) audio, (\textit{b}) text, (\textit{c}) visual, (\textit{d}) all modalities. We show positive (\textit{blue}) and negative samples (\textit{red}). 
		}
		\label{fig:vis_feat}
\end{figure}


\subsection{Modality Impact} 
\label{sub:analysis_modality}

To visualize the impact of modalities, we compute the average attention values on the three CA modules (CA boxes in Figure~\ref{fig:main}) and then, show the average weights for each modality in the pie chart of each example in Figure~\ref{fig:vis_cls}. For this, we show (a-d) four positive and (e,f) two negative samples on Friends with frames, subtitles and audio spectrogram (left) and pitch (right). We observe that the contribution of each modality varies; the commonality though is that audio contributes more than half, followed by text and finally visual features. Specifically, in cases where there is a strong audio signal, the contribution of audio increases significantly. This is illustrated when the character yells (`Chandler' in positive example (a), or pauses the speech (`Chandler' in positive example (b), or the speech rate speeds up (`Phoebe' in positive example (c) or speech volumes change suddenly (`Chandler' and `Joey' in positive example (d). In contrast, in negative (e) and (f), the tone, volume, pitch or rhythm do not change greatly, so the text starts to play a bigger role in determining them as non-funny scenes. Furthermore, we observe that in the (c) and (e) examples, the visual feature plays very little role in the final prediction probably because the scenes do not capture the whole character's bodies and their movement, so the visual model can offer only little information.

\subsection{Feature Visualization} 
\label{sub:analysis_visualization}

Figure~\ref{fig:vis_feat} shows the t-SNE~\cite{t-sne} visualization of features: (a, b, c) display the unimodal distributions of audio, text, and visual features respectively, while (d) corresponds to all modalities for four datasets. Blue colour corresponds to funny samples and red to not-funny ones. All single features, and in particular the visual and textual ones, are scattered in the 2D space without clear boundaries between positives and negatives. Interestingly, for Friends, TBBT and MUSTaRD the audio features alone exhibit a notable ability to discriminate positive and negative samples; this is probably because of the punchlines used in these shows that typically occur at the end of sentences. 
For these three datasets, we observe that the joint embedding of all modalities results in the best separation between positives and negatives. 
Interestingly, for UR-Funny, a dataset without ending punchlines, all combinations of modalities (either single or joint) fail to distinguish funny from not-funny moments. This is probably due to the domain shift between samples from this dataset (TED-talk segments) and the samples used at training (sitcoms).

\begin{figure}[t]

\centerline{\includegraphics[width=\columnwidth]{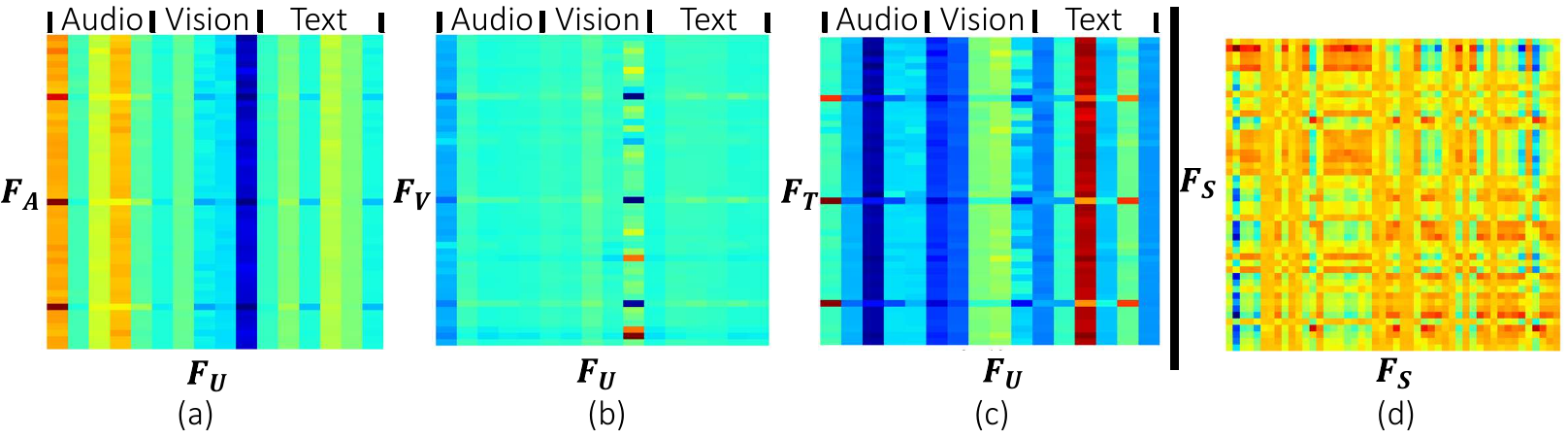} }
		\caption{
		CAF attention maps on the test set of Friends.
		(\textit{a},\textit{b},\textit{c}) Cross Attention between the unified feature $F_{U}$ (coming from all modalities) and audio, vision and text; (\textit{d}) Self Attention on $F_{U}$.
  } 
  
		\label{fig:caf_visual}
\end{figure}

\begin{figure*}[t]
\includegraphics[width=\linewidth]{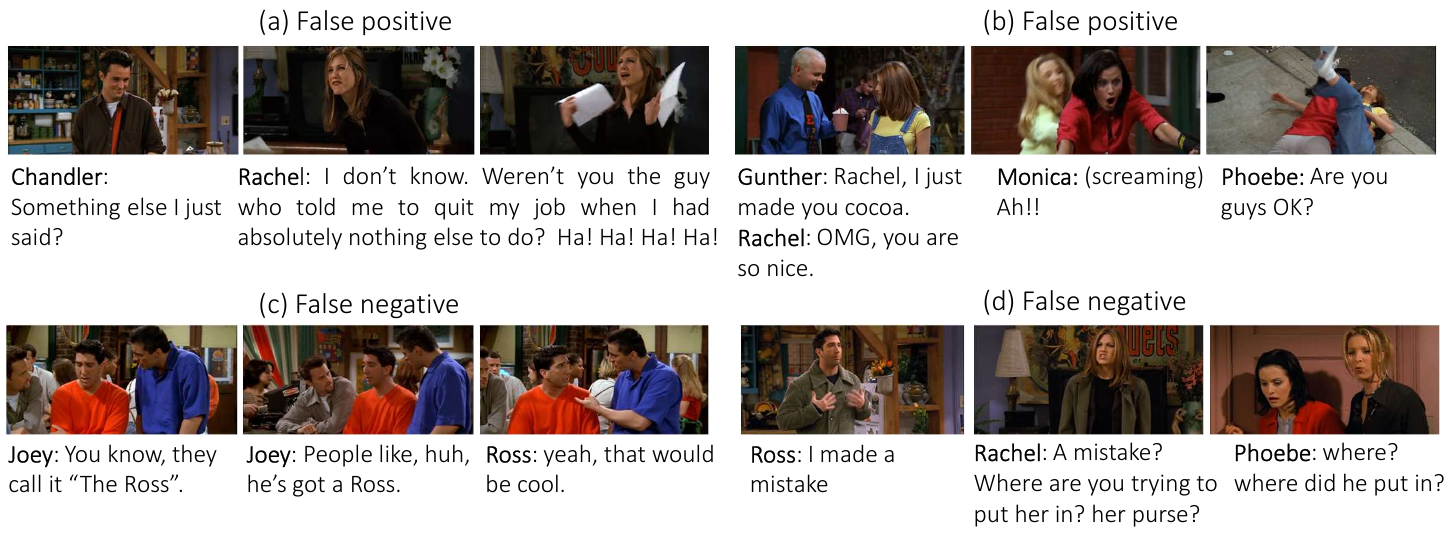}
    \caption{
    Failure cases on Friends split into three main groups: (a, b) strong emotional responses expressed by single wording, (c) subtle sarcastic comments with straight face and no follow-up indications, and (d) inside jokes depending on long-term understanding. 
    }
\label{fig:vis_failure}
\end{figure*}

\subsection{Impact of CAF Module}
\label{sub:analysis_caf}

To examine the effect of CAF, we visualize in Figure~\ref{fig:caf_visual} the learned attention maps: red indicates higher and blue lower attention. 
(a,b,c) display the cross-attention between the unified $F_{\text{U}}$ 
and (a) audio, (b) visual, (c) text features.  
Since $F_{\text{U}}$ is stacked from audio, vision, and text, we observe that each modality highly attends to itself (especially text). We also observe that the audio encoder also attends to the text encoder, indicating that there is mutual information shared between text and audio. Finally, (d) displays the self-attention map between $F_{\text{U}}$, where we observe that $F_{\text{U}}$ attends to all tokens with different weights. The small color differences on the diagonal and anti-diagonal areas suggest that the joint features have approximately uniform representations for the final classification.

\subsection{Failure Analysis}
\label{sub:analysis_failure}

By examining the results, we observe three main groups of failure cases. 
First, when characters have strong emotional responses expressed only by single words (such as `haha', `no!') is not always funny. However, all modalities incorrectly, yet confidently predict them as funny. Figures~\ref{fig:vis_failure}-(a) and (b) show this. In Figure~\ref{fig:vis_failure}-(a), `Rachel' laughs sarcastically, which is not funny (subtitle `ha ha'). FunnyNet-W incorrectly predicts it as positive. In Figure~\ref{fig:vis_failure}-(b), Monica screams loudly and falls on Rachel (subtitle `Ah'). The sudden high pitch of the scream gives the wrong signal to the model and wrongly predicts it as positive.   
Second, when the funny moment is expressed only with subtle indications, typically sarcasm without a follow-up signal (indicative facial expression or grimace, surprise, pause in dialogue, phrase, joke). In such cases, FunnyNet-W may fail to discriminate these subtle cues that come usually with human-level understanding. 
Figure~\ref{fig:vis_failure}-(c) indicates such an example, where Ross gives a sarcastic response to Joey without changing facial expression or tone; in this case, FunnyNet-W incorrectly predicts the scene as negative. 
Third, in most cases, all modalities fail to understand inside jokes that depend on long-term dependencies.
For instance, Figure~\ref{fig:vis_failure}-(d) is the case where the context is so long (the previous awkward moment between Ross and Rachel) that the model wrongly predicts the scene as a not-funny moment. All audio, visual or text fail to give discriminative signals to indicate the funniness.

\begin{figure*}[t]
\includegraphics[width=\linewidth]
{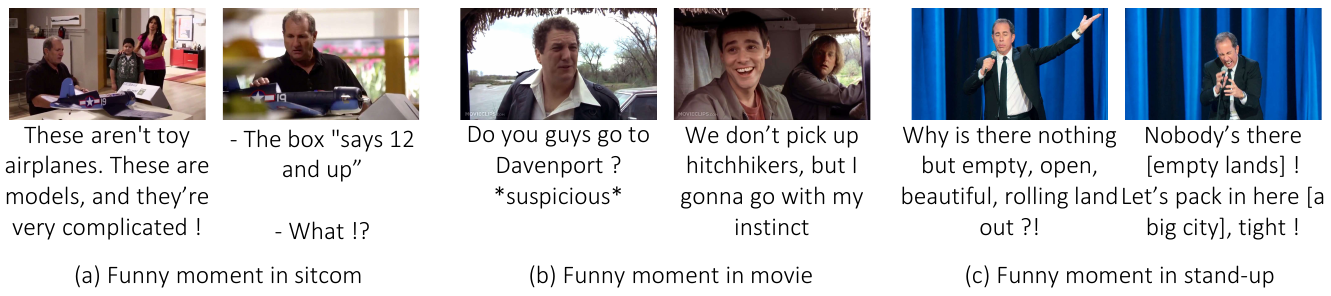} 
		\caption{Funny moments in the wild. Three examples from sitcoms, movies, and stand-up with diverse contents, e.g. the sitcom does not have canned laughter, the movie contains dramatic acting performance, and the stand-up is a one-man show without interactions.}
		\label{fig:coolappli1}
\end{figure*}

\section{Funny Scene Detection in the Wild}
\label{sub:wild}

\subsection{Applications from other domains}
\label{sub:coolappli}

In this section, we show applications of FunnyNet and FunnyNet-W in videos from other domains. 

\paragraph{1. Sitcoms without Canned Laughter.} 
In FunnyNet~\cite{funnynet}, we collect 9 episodes of the first season ($\sim$180 minutes) of \textit{Modern Family} (Lloyd and Levitan, 2009)\footnote{\url{https://www.youtube.com/playlist?list=PL8v3aNB88WMM0iwOUeLpgFf3pHH9uxz7_}.} without canned laughter. We manually annotate as positive every punchline that could lead to laughter, resulting in $453$ positives (we will make them available). 
Figure~\ref{fig:coolappli1}~(a) shows a correctly predicted funny moment between two characters who vary their speech rhythm and tones.

\paragraph{2. Movies with Diverse Funny Styles.} Figure~\ref{fig:coolappli1}~(b) depicts such an example from the \textit{Dumb and Dumber} film (Farrelly, 1994). Our model correctly detects funny moments followed by silence or a speaker's change of tone. 

\paragraph{3. Stand-Up Comedies.} They contain several punchlines that make audiences laugh. We experiment on the Jerry Seinfeld \textit{23 Hours to Kill} stand-up comedy. Figure~\ref{fig:coolappli1}~(c) shows that FunnyNet detects funny moments correctly and confidently, as Jerry is highly expressive (expressions, gestures). 

\paragraph{4. Audio-Only.} As audio is the most discriminative cue, we examine its impact on out-of-domain audios: narrating jokes and reading books. Our model detects funny punchlines from jokes, mostly when they are accompanied by a change of pitch or pause; for the audiobook, it successfully detects funny moments when the reader's voice imitates a character.

\subsection{FunnyNet-W against LLM Chatbot}
\label{sub:wild_chatbot}

Recently, several large language models (LLMs)~\cite{openai2023gpt4,touvron2023llama} have been fine-tuned and are used as chatbots~\cite{chatgpt,köpf2023openassistant}. With their expansive knowledge and context-aware responses, they have significantly advanced language understanding and generation, which enable them to perform a wide range of language-related tasks. 
In this context, we compare the proposed FunnyNet-W against a chatbot to assess its performance relative to these general models. Specifically, we use the LlaMa-2~\cite{llama2} chatbot on the Friends dataset. 

\paragraph{Prompting.}
We evaluate the language LlaMa-2 chatbot in two setups. 

\noindent First, with or without prompt training:
\begin{itemize}
    \item \textbf{zero-shot setting}, where we prompt the chatbot with a transcript sample and ask it to determine whether it is funny or not. We do that iteratively for all test samples of the Friends test set. 
    The prompt we use is ``\textit{Is the following sentence funny or not?} $\ll$subtitles$\gg$'', where $\ll$subtitles$\gg$ corresponds to each test sample. 
    However, given the popular nature of the `Friends' sitcom,  the chatbot may have already seen samples or even the whole transcript of the TV show during training. We hypothesize that this impacts its performance positively, as the chatbot not only has knowledge of the dialogues that follow, but also knows the comments of the community for each pun or joke. 
    \item \textbf{few-shot setting}, where we prompt the chatbot with some training samples followed by the testing sample within the token context limit. The prompt we use is twenty training samples (ten positives and ten negatives): ``\textit{This sentence is funny:} $\ll$subtitles$\gg$. \textit{This sentence is not funny} $\ll$subtitles$\gg$.'', followed by the testing sample: ``\textit{Is the following sentence funny or not?} $\ll$subtitles$\gg$''. In this case, the chatbot uses the training samples to better distinguish the specific TV show type of humour.
\end{itemize}

Second, by performing a simple prompt engineering (i.e. part of the prompt that gives context to the chatbot):
\begin{itemize}
    \item \textbf{general system prompt}, we prompt the chatbot with the general system prompt (referred to as `Generic'): ``\textit{You are a helpful, respectful and honest assistant. Always answer as helpfully as possible, while being safe.  Your answers should not include any harmful, unethical, racist, sexist, toxic, dangerous, or illegal content. [...]}''. This system prompt makes the chatbot act as a general chatbot without any prior on the task. 
    \item \textbf{specific system prompt}, we prompt the chatbot with the task-specific system prompt (referred to as `Specific'): ``\textit{I will give some sentences, and you need to say if it’s funny or not, reply only by yes or no.}''. This kind of system prompt helps the chatbot limit its range and focus on the task only. Moreover, it forces the chatbot to answer, whereas the general system prompt leads sometimes to hesitating answers. 
\end{itemize}

\begin{table}[t]
    \caption{Chatbot vs FunnyNet-W.} 
    \centering
        \begin{small}
    \resizebox{\linewidth}{!}{
\begin{tabular}{cc|cc}
\toprule
Prompt engineering & Prompt training & F1 & Accuracy \\
\midrule
\multirow{2}{*}{Generic} & - & 14.5 &	41.8 \\
 & \checkmark & 44.3 & 46.5 \\
\multirow{2}{*}{Specific} & - & 64.1 & 53.2 \\
 & \checkmark & 71.1 & 55.9 \\
\midrule
\multicolumn{2}{c|}{FunnyNet-W (T)} & 77.8 & 68.1 \\
\multicolumn{2}{c|}{FunnyNet-W (A+V+T)} & 88.2 & 85.6 \\
\bottomrule
\end{tabular}}
\end{small}
\label{tab:chatbot}
\end{table}

\paragraph{Experimental results.}
Table~\ref{tab:chatbot} reports the results when prompting the  LlaMa-2 chatbot. 
We observe that without prompt training, the chatbot's performance drops both with and without prompt engineering. Additionally, we observe the importance of prompt engineering: when using the specific prompt (with or without training) the performances are higher than 50\% in both metrics, whereas the generic prompt (no prompt engineering) results in very low performances. 
This is in line with the current bibliography on LLMs, where prompt engineering is crucial for higher accuracy. We believe that more prompt engineering will increase the performance; yet, this is outside the scope of this work. Overall, we observe that FunnyNet-W outperforms all examined cases with the chatbot. This highlights the need for specific model training for funny moment detection. Interestingly, we note that the performance of FunnyNet-W using text only is close to the one of the LlaMa-2 chatbot, thus showcasing the impressive representation power of LLM chatbots. 

\begin{figure}[t]
\centering
		\centerline{\includegraphics[width=\columnwidth]{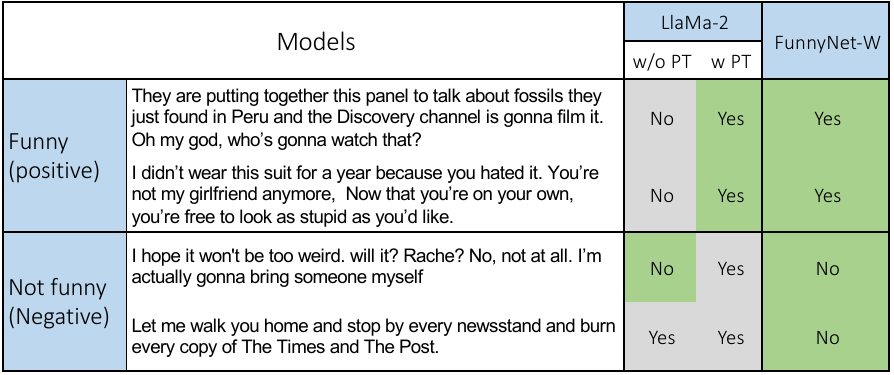}} 
		\caption{
		Examples of using Chatbot, with or without prompt training (w PT or w/o PT), and FunnyNet-W for funny prediction. FunnyNet-W can give correct predictions (green) for both positive and negative examples, while LlaMa-2 fails to give good results and the prompt training brings small improvements.
		}
		\label{fig:text_example}
\end{figure}

Figure~\ref{fig:text_example} illustrates four examples: two positive (first and second rows) and two negative samples (third and fourth rows). For the positive samples, we observe that LlaMa-2 correctly understand its funniness (green) with prompt training, whereas when there is no prompt the results are incorrect. For two negative samples, most predictions from the chatbot are incorrect, most likely because it picks up the words with strong emotional expressions, like ``weird'' and ``burn'', resulting in false positives. In all examples, FunnyNet-W correctly predicts the results because it does not solely rely on text, but also audio and visual features.

Overall, our findings are twofold: (1) using only subtitles is insufficient to understand the funniness in video scenes, and (2) since we only do minor prompt engineering (generic and specific), the results of LLMs cannot outperform the proposed FunnyNet-W. Potentially, by improving the prompts, we can further improve the performance of LLMs. 

\subsection{Impact of Audio}
\label{sub:wild_audio}

\begin{table}[t]
    \caption{Ablation of synthetic and real voice when training and testing FunnyNet-W on Friends.}

    \centering
        \begin{small}
        \resizebox{\linewidth}{!}{
            \begin{tabular}{cc|cccc}
            \toprule
            \multicolumn{2}{c|}{\multirow{3}{*}{Model}} & \multicolumn{4}{c}{Training voice} \\
            \multicolumn{2}{c|}{} & \multicolumn{2}{c}{synthetic} & \multicolumn{2}{c}{real} \\
            \multicolumn{2}{c|}{} & F1 & Accuracy & F1 & Accuracy \\
            \midrule
            \multirow{2}{*}{Test voice} & synthetic & 65.5 & 67.7 & 68.8 & 66.8 \\
             & real & 83.1 & 79.6 & 88.2 & 85.6 \\
             \bottomrule
            \end{tabular}}
        \end{small}
\label{tab:voice}
\end{table}

In the context of funny moment detection, audio is more relevant than text~\cite{funnynet} because it contains more information, including vocals, pauses, pitch variations, speech rate variations, rhythm and timing, accent and pronunciation, emotional tone, music and background noise. To highlight the importance of audio, in this section, we test FunnyNet-W by replacing the ground truth audio with automatic machine sounds. 

For this, we generate corresponding synthetic audios from the ground truth subtitles of the Friends dataset with a text-to-audio model\footnote{\url{https://github.com/pndurette/gTTS}}. Note that the synthetic audio only mimics the vocals between characters without any background sounds and, more importantly, without including all additional voice cues that help identify the emotional state of the character and the dialogue.  Then, we train FunnyNet-W with the synthetic voices and test it on both the real and synthetic voices and respectively, we test FunnyNet-W (trained on real voices) on both real and synthetic voices.  Table~\ref{tab:voice} reports the results. 
When training with synthetic voice (first and second column), we observe that testing on real voices (second row) outperforms testing on synthetic ones (first row) by a large margin, i.e. approximately 10-15\% for both metrics. Similarly, when training with real voice (third and fourth columns), we observe that there is a significant difference in performance (or approximately 20\% in both metrics) between testing on synthetic and real data. These results show that simply replacing real voice with synthetic ones omits other important information, such as background audio, and music; hence, the model makes more correct predictions when the test set contains additional auditory information (real) rather than a simple voice (synthetic).   
When we test on synthetic voices (first row), we observe that 
training either with synthetic or real voice produces similar results. This is because the test set contains synthetic data, and therefore learning the specificities of voice is not necessary for good performance. 
However, when we test on real voices (second row), we observe that training with real voices (columns 3-4) outperforms training with synthetic ones (columns 1-2) by a large margin (e.g. for Acc 79.6\% for synthetic vs. 85.6\% for real). This clearly shows that the real voice includes important additional cues (pause, intonation, etc.) that help FunnyNet-W discriminate funniness. 

\begin{figure}[t]
\centering
	\centerline{\includegraphics[width=\columnwidth]{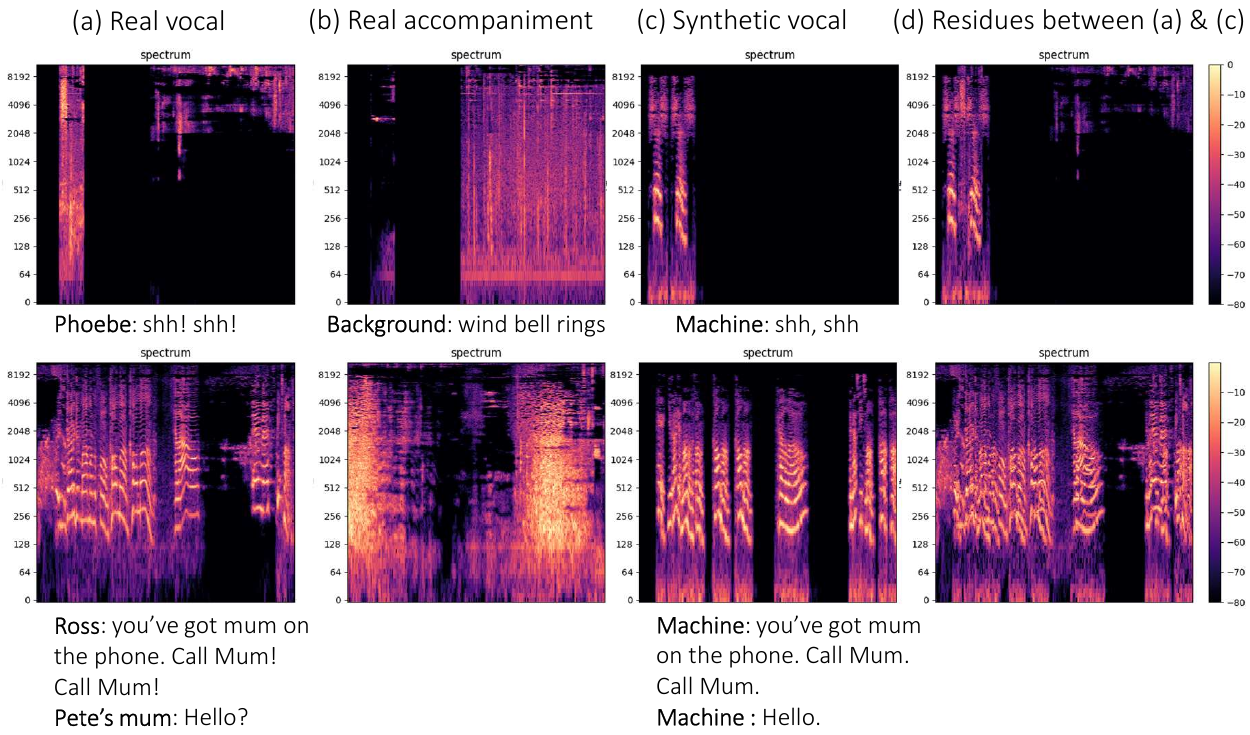}} 
		\caption{\small{ 
		Visualization of real and synthetic audio on Friends. We show real vocals (\textit{a}), real accompaniment (\textit{b}), synthetic vocals (\textit{c}) and the residues between real and synthetic vocals (\textit{d}). 
		}}
		\label{fig:audio_abla}
\end{figure}

\begin{figure}[t]
\centering
	\centerline{\includegraphics[width=\columnwidth]{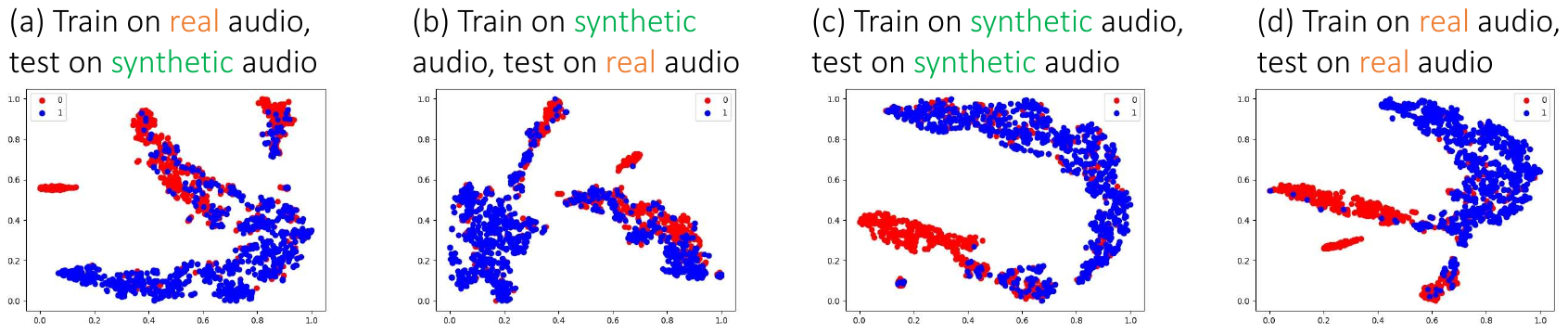}} 
		\caption{\small{ 
		T-SNE visualization of real and synthetic audio on Friends. We show positive (\textit{blue}) and negative (\textit{red}) samples to indicate the feature distributions. 
		}}
		\label{fig:audio_tsne}
\end{figure}

To further analyze the effect of voice, we perform here a qualitative comparison using Spleeter\footnote{https://github.com/deezer/spleeter}. Specifically, Figure~\ref{fig:audio_abla} illustrates the spectrum heatmaps between (a) real vocal, (b) real accompaniment (non-vocal parts, such as background music, sounds, talks, audio), (c) synthetic vocal, and (d) the differences between real and synthetic audio. 
We visualize the heatmaps of examples (two rows), where in both cases FunnyNet-W correctly predicts the funniness when using real audio and incorrectly when using synthetic audio. The first row shows the funny moment when Phoebe tries to shush the wind bell while the wind bell keeps ringing. This contrast between vocals and non-vocal sounds (i.e. in this case bell ringing) is missing from the synthetic vocals. Row 2 shows that when Ross screams excitedly (`Call Mum!'), his voice triggers the smartphone to dial Pete's mum. This strong vocal expression does not appear on the synthetic vocals. Both these examples indicate that audio plays a key role in funny moment detection because it contains not only background sounds but also expressions and feelings from the characters leading to better scene understanding. 

Furthermore, we also use T-SNE to visualize the data clustering in Figure~\ref{fig:audio_tsne}. This visualization shows that when the train and test data come from the same domain (either real or synthetic, c and d figures), the positive and negative distributions (blue and red points) are clearly separable. This is in contrast to the (a,b) figures, where the two synthetic and real domains are mixed (i.e., training on one domain and testing on another); in this case, the two distributions overlap more, as expected due to domain shift~\cite{kalogeiton2016analysing,torralba2011unbiased}.

\section{Ethical discussion}

\paragraph{Practical Impact.}
There are various potential applications for FunnyNet-W. 
First, it may be useful to collect a large dataset of funny moments (similar to~\cite{li2023oxfordtvghic}), so for example, cognitive researchers could study funniness mechanisms at a large scale. 
Next, it may be useful to enable artists to edit films more easily, without relying on a live audience. 
Finally, it may be useful to enhance human-machine interactions. For instance, adding a sense of humor to conversational agents would make the relation more natural and spontaneous.

However, FunnyNet-W is part of artificial intelligence systems that tend to analyze complex human specificities and behaviors (e.g., conversational agents). Given the nature of these systems, their usage and deployment should be done with caution. 
For instance, in the particular case of FunnyNet-W, it could enhance identity fraud methods, by better mimicking the sense of humor of victims.

\paragraph{Societal Impact.}
FunnyNet-W is trained mainly with Western cultural materials, especially from the USA, which do not necessarily represent uniform demographics. 
In particular, we mainly tackle funniness in American sitcoms, which covers a very specific type of humor.
Therefore, without fine-tuning, FunnyNet-W might have difficulties in generalizing to funny moments from other cultures, as humor is highly thematic, and themes vary from one culture to another. 
Moreover, the audio modality might also be highly impacted by cultural bias, as expressiveness is strongly related to culture, e.g., actor performances change a lot from one country to another, leading to misinterpretations.
In addition to the cultural barrier, FunnyNet-W includes language bias. 
Indeed, the audio as well as the textual modality are trained with the English language. This can be a limiting factor for generalization and transferability across languages, as jokes or puns often rely on language specificities. We also note that the textual modality is limited by alphabets that vary among languages.

\paragraph{Environmental Impact.}
All experiments are done on NVIDIA RTX4090 and A100 GPUs, with each of them requiring 215W in power supply. 
For this project, we use approximately 800 GPU hours. 
Training a FunnyNet-W model with all three modalities requires around 6 GPU hours on NVIDIA RTX4090, which amounts to 1.29 kWh and 300.75g of CO2 emitted.

\section{Conclusions}
\label{sec:conclusions}
We introduced FunnyNet, an audiovisual model for funny moment detection. 
In contrast to works that rely on text, FunnyNet exploits audio that comes naturally with videos and contains high-level cues (pauses, tones, etc). 
Our findings show audio is the dominant cue for signaling funny situations, while video offers complementary information. 
Extensive analysis and visualizations also support our finding that audio is better than text (in the form of subtitles) when it comes to scenes with no or simple dialogue but with hilarious acting or funny background sounds.  
Our results show the effectiveness of each component of FunnyNet, which outperforms the state of the art on the TBBT, MUStARD, MHD, UR-Funny and Friends. Future work includes analyzing the contribution of audio cues (pitch, tone, etc).

\vspace{5mm}
\noindent \textbf{Acknowledgements.} 
This work is supported by a DIM RFSI grant, a Hi!Paris collaborative project grant, the ANR projects WhyBehindScenes ANR-22-CE23-0007 and APATE ANR-22-CE39-0016, and the HPC resources of IDRIS under the allocation 2022-AD011013951 made by GENCI.

\bibliographystyle{splncs04}
\bibliography{shortstrings,references}

\begin{thebibliography}{100}
\providecommand{\url}[1]{\texttt{#1}}
\providecommand{\urlprefix}{URL }
\providecommand{\doi}[1]{https://doi.org/#1}

\bibitem{audio_vision_2}
Afouras, T., Chung, J.S., Zisserman, A.: The conversation: Deep audio-visual speech enhancement. In: INTERSPEECH (2020)

\bibitem{colbert}
Annamoradnejad, I., Zoghi, G.: Colbert: Using bert sentence embedding for humor detection. arXiv preprint arXiv:2004.12765  (2020)

\bibitem{bain2022whisperx}
Bain, M., Huh, J., Han, T., Zisserman, A.: Whisperx: Time-accurate speech transcription of long-form audio. arXiv preprint, arXiv:2303.00747  (2023)

\bibitem{bain2021frozen}
Bain, M., Nagrani, A., Varol, G., Zisserman, A.: Frozen in time: A joint video and image encoder for end-to-end retrieval. In: ICCV (2021)

\bibitem{barral2017no}
Barral, O., Kosunen, I., Jacucci, G.: No need to laugh out loud: Predicting humor appraisal of comic strips based on physiological signals in a realistic environment. ACM Transactions on Computer-Human Interaction  (2017)

\bibitem{timesformer}
Bertasius, G., Wang, H., Torresani, L.: Is space-time attention all you need for video understanding? In: ICML (2021)

\bibitem{audio_humor}
Bertero, D., Fung, P.: Deep learning of audio and language features for humor prediction. In: LREC (2016)

\bibitem{brown2021face}
Brown, A., Kalogeiton, V., Zisserman, A.: Face, body, voice: Video person-clustering with multiple modalities. In: ICCV (2021)

\bibitem{mustard}
Castro, S., Hazarika, D., P{\'e}rez-Rosas, V., Zimmermann, R., Mihalcea, R., Poria, S.: Towards multimodal sarcasm detection (an \textit{{O}bviously} perfect paper). In: ACL (2019)

\bibitem{chen2022beats}
Chen, S., Wu, Y., Wang, C., Liu, S., Tompkins, D., Chen, Z., Wei, F.: Beats: Audio pre-training with acoustic tokenizers. ICML  (2023)

\bibitem{simclr}
Chen, T., Kornblith, S., Norouzi, M., Hinton, G.: A simple framework for contrastive learning of visual representations. In: ICML (2020)

\bibitem{chung2016out}
Chung, J.S., Zisserman, A.: Out of time: automated lip sync in the wild. In: ACCV (2016)

\bibitem{chung2019perfect}
Chung, S.W., Chung, J.S., Kang, H.G.: Perfect match: Improved cross-modal embeddings for audio-visual synchronisation. In: ICASSP (2019)

\bibitem{sarcasm_1}
Davidov, D., Tsur, O., Rappoport, A.: Semi-supervised recognition of sarcastic sentences in twitter and amazon. In: ACL (2010)

\bibitem{demucs}
D{\'e}fossez, A., Usunier, N., Bottou, L., Bach, F.: Music source separation in the waveform domain. arXiv preprint arXiv:1911.13254  (2019)

\bibitem{deng2018multimodal}
Deng, D., Zhou, Y., Pi, J., Shi, B.E.: Multimodal utterance-level affect analysis using visual, audio and text features. arXiv preprint arXiv:1805.00625  (2018)

\bibitem{bert}
Devlin, J., Chang, M.W., Lee, K., Toutanova, K.: Bert: Pre-training of deep bidirectional transformers for language understanding. In: NAACL (2019)

\bibitem{dong2021dual}
Dong, J., Li, X., Xu, C., Yang, X., Yang, G., Wang, X., Wang, M.: Dual encoding for video retrieval by text. IEEE TPAMI  (2021)

\bibitem{dufour22eccv}
Dufour, N., Picard, D., Kalogeiton, V.: Scam! transferring humans between images with semantic cross attention modulation. In: ECCV (2022)

\bibitem{epstein2021learning}
Epstein, D., Vondrick, C.: Learning goals from failure. In: CVPR (2021)

\bibitem{fang2021clip2video}
Fang, H., Xiong, P., Xu, L., Chen, Y.: Clip2video: Mastering video-text retrieval via image clip. arXiv preprint arXiv:2106.11097  (2021)

\bibitem{audio_vision_1}
Gabbay, A., Ephrat, A., Halperin, T., Peleg, S.: Seeing through noise: Visually driven speaker separation and enhancement. In: ICASSP (2018)

\bibitem{gabeur2020multi}
Gabeur, V., Sun, C., Alahari, K., Schmid, C.: Multi-modal transformer for video retrieval. In: ECCV (2020)

\bibitem{gemmeke2017audio}
Gemmeke, J.F., Ellis, D.P., Freedman, D., Jansen, A., Lawrence, W., Moore, R.C., Plakal, M., Ritter, M.: Audio set: An ontology and human-labeled dataset for audio events. In: ICASSP (2017)

\bibitem{new_laughter_detector}
Gillick, J., Deng, W., Ryokai, K., Bamman, D.: Robust laughter detection in noisy environments. INTERSPEECH  (2021)

\bibitem{girdhar2023imagebind}
Girdhar, R., El-Nouby, A., Liu, Z., Singh, M., Alwala, K.V., Joulin, A., Misra, I.: Imagebind: One embedding space to bind them all. arXiv preprint arXiv:2305.05665  (2023)

\bibitem{gong2023cavmae}
Gong, Y., Rouditchenko, A., Liu, A.H., Harwath, D., Karlinsky, L., Kuehne, H., Glass, J.R.: Contrastive audio-visual masked autoencoder. In: ICLR (2023)

\bibitem{guzhov2022audioclip}
Guzhov, A., Raue, F., Hees, J., Dengel, A.: Audioclip: Extending clip to image, text and audio. In: ICASSP (2022)

\bibitem{autoad}
Han, T., Bain, M., Nagrani, A., Varol, G., Xie, W., Zisserman, A.: Autoad ii: The sequel - who, when, and what in movie audio description. In: ICCV (2023)

\bibitem{hasan2021humor}
Hasan, M.K., Lee, S., Rahman, W., Zadeh, A., Mihalcea, R., Morency, L.P., Hoque, E.: Humor knowledge enriched transformer for understanding multimodal humor. In: AAAI (2021)

\bibitem{urfunny}
Hasan, M.K., Rahman, W., Bagher~Zadeh, A., Zhong, J., Tanveer, M.I., Morency, L.P., Hoque, M.E.: {UR}-{FUNNY}: A multimodal language dataset for understanding humor. In: EMNLP-IJCNLP (2019)

\bibitem{MISA}
Hazarika, D., Zimmermann, R., Poria, S.: Misa: Modality-invariant and-specific representations for multimodal sentiment analysis. ACM International Conference on Multimedia  (2020)

\bibitem{t-sne}
Hinton, G., Roweis, S.: Stochastic neighbor embedding. In: NeurIPS (2002)

\bibitem{switchboard}
Holliman, E., Godfrey, J., McDaniel, J.: Switchboard: telephone speech corpus for research and development. In: ICASSP (1992)

\bibitem{hyperbolic}
Hong, J., Hayder, Z., Han, J., Fang, P., Harandi, M., Petersson, L.: Hyperbolic audio-visual zero-shot learning. In: ICCV (2023)

\bibitem{huber2012modern}
Huber, D.M., Runstein, R.: Modern recording techniques. Routledge (2012)

\bibitem{iashin2020multi}
Iashin, V., Rahtu, E.: Multi-modal dense video captioning. In: CVPR-workshops (2020)

\bibitem{jaegle2021perceiver}
Jaegle, A., Gimeno, F., Brock, A., Zisserman, A., Vinyals, O., Carreira, J.: Perceiver: General perception with iterative attention. ICML  (2021)

\bibitem{kalogeiton2016analysing}
Kalogeiton, V., Ferrari, V., Schmid, C.: Analysing domain shift factors between videos and images for object detection. IEEE TPAMI  (2016)

\bibitem{kalogeiton2020constrained}
Kalogeiton, V., Zisserman, A.: Constrained video face clustering using 1nn relations. In: BMVC (2020)

\bibitem{laught_machine}
Kayatani, Y., Yang, Z., Otani, M., Garcia, N., Chu, C., Nakashima, Y., Takemura, H.: The laughing machine: Predicting humor in video. In: WACV (2021)

\bibitem{kim2023prefix}
Kim, M., Sung-Bin, K., Oh, T.H.: Prefix tuning for automated audio captioning. In: ICASSP (2023)

\bibitem{koepke2022audio}
Koepke, A.S., Oncescu, A.M., Henriques, J., Akata, Z., Albanie, S.: Audio retrieval with natural language queries: A benchmark study. IEEE Transactions on Multimedia  (2022)

\bibitem{koizumi2020transformer}
Koizumi, Y., Masumura, R., Nishida, K., Yasuda, M., Saito, S.: A transformer-based audio captioning model with keyword estimation. arXiv preprint arXiv:2007.00222  (2020)

\bibitem{korbar2018co}
Korbar, B.: Co-training of audio and video representations from self-supervised temporal synchronization. In: CoRR (2018)

\bibitem{köpf2023openassistant}
Köpf, A., Kilcher, Y., von Rütte, D., Anagnostidis, S., Tam, Z.R., Stevens, K., Barhoum, A., Duc, N.M., Stanley, O., Nagyfi, R., ES, S., Suri, S., Glushkov, D., Dantuluri, A., Maguire, A., Schuhmann, C., Nguyen, H., Mattick, A.: Openassistant conversations -- democratizing large language model alignment. arXiv preprint arXiv:2304.07327  (2023)

\bibitem{lee2020cross}
Lee, J.T., Jain, M., Park, H., Yun, S.: Cross-attentional audio-visual fusion for weakly-supervised action localization. In: ICLR (2020)

\bibitem{li2023oxfordtvghic}
Li, R., Sun, S., Elhoseiny, M., Torr, P.: Oxfordtvg-hic: Can machine make humorous captions from images? arXiv preprint arXiv:2307.11636  (2023)

\bibitem{liang2020learning}
Liang, Z., Jiang, W., Hu, H., Zhu, J.: Learning to contrast the counterfactual samples for robust visual question answering. In: EMNLP (2020)

\bibitem{lin2022swinbert}
Lin, K., Li, L., Lin, C.C., Ahmed, F., Gan, Z., Liu, Z., Lu, Y., Wang, L.: Swinbert: End-to-end transformers with sparse attention for video captioning. In: CVPR (2022)

\bibitem{lin2022frozen}
Lin, Z., Geng, S., Zhang, R., Gao, P., de~Melo, G., Wang, X., Dai, J., Qiao, Y., Li, H.: Frozen clip models are efficient video learners. In: ECCV (2022)

\bibitem{liu2022visually}
Liu, X., Huang, Q., Mei, X., Liu, H., Kong, Q., Sun, J., Li, S., Ko, T., Zhang, Y., Tang, L.H., et~al.: Visually-aware audio captioning with adaptive audio-visual attention. INTERSPEECH  (2023)

\bibitem{funnynet}
Liu, Z.S., Courant, R., Kalogeiton, V.: Funnynet: Audiovisual learning of funny moments in videos. In: ACCV (2022)

\bibitem{adamw}
Loshchilov, I., Hutter, F.: Decoupled weight decay regularization. In: ICLR (2019)

\bibitem{lou2022audio}
Lou, S., Xu, X., Wu, M., Yu, K.: Audio-text retrieval in context. In: ICASSP (2022)

\bibitem{mei2022language}
Mei, X., Liu, X., Liu, H., Sun, J., Plumbley, M.D., Wang, W.: Language-based audio retrieval with pre-trained models. DCASE  (2022)

\bibitem{mesaros2017detection}
Mesaros, A., Heittola, T., Benetos, E., Foster, P., Lagrange, M., Virtanen, T., Plumbley, M.D.: Detection and classification of acoustic scenes and events: Outcome of the dcase 2016 challenge. IEEE/ACM Transactions on Audio, Speech, and Language Processing  (2017)

\bibitem{mesaros2016tut}
Mesaros, A., Heittola, T., Virtanen, T.: Tut database for acoustic scene classification and sound event detection. In: European Signal Processing Conference (2016)

\bibitem{cross_attn_1}
Mohla, S., Pande, S., Banerjee, B., Chaudhuri, S.: Fusatnet: Dual attention based spectrospatial multimodal fusion network for hyperspectral and lidar classification. In: CVPR (2020)

\bibitem{morgado2021audio}
Morgado, P., Vasconcelos, N., Misra, I.: Audio-visual instance discrimination with cross-modal agreement. In: CVPR (2021)

\bibitem{nagrani2021attention}
Nagrani, A., Yang, S., Arnab, A., Jansen, A., Schmid, C., Sun, C.: Attention bottlenecks for multimodal fusion. In: NeurIPS (2021)

\bibitem{cross_attn_3}
Nam, H., Ha, J.W., Kim, J.: Dual attention networks for multimodal reasoning and matching. In: CVPR (2017)

\bibitem{narasimhan2021clip}
Narasimhan, M., Rohrbach, A., Darrell, T.: Clip-it! language-guided video summarization. In: NeurIPS (2021)

\bibitem{byol_a}
Niizumi, D., Takeuchi, D., Ohishi, Y., Harada, N., Kashino, K.: Byol for audio: Self-supervised learning for general-purpose audio representation. In: International Joint Conference on Neural Networks (2021)

\bibitem{byola_v2}
Niizumi, D., Takeuchi, D., Ohishi, Y., Harada, N., Kashino, K.: Byol for audio: Exploring pre-trained general-purpose audio representations. IEEE/ACM Transactions on Audio, Speech, and Language Processing  (2023)

\bibitem{oord2018representation}
Oord, A.v.d., Li, Y., Vinyals, O.: Representation learning with contrastive predictive coding. arXiv preprint arXiv:1807.03748  (2018)

\bibitem{chatgpt}
OpenAI: {ChatGPT}: Conversational ai powered by {GPT}-3.5. OpenAI Blog  (2021)

\bibitem{openai2023gpt4}
OpenAI: Gpt-4 technical report. arXiv preprint arXiv:2303.08774  (2023)

\bibitem{owens2018audio}
Owens, A., Efros, A.A.: Audio-visual scene analysis with self-supervised multisensory features. In: ECCV (2018)

\bibitem{NEURIPS2019_9015}
Paszke, A., Gross, S., Massa, F., Lerer, A., Bradbury, J., Chanan, G., Killeen, T., Lin, Z., Gimelshein, N., Antiga, L., et~al.: Pytorch: An imperative style, high-performance deep learning library. In: NeurIPS (2019)

\bibitem{humor_data}
Patro, B.N., Lunayach, M., Srivastava, D., Sarvesh, S., Singh, H., Namboodiri, V.P.: Multimodal humor dataset: Predicting laughter tracks for sitcoms. In: WACV (2021)

\bibitem{priyasad2020attention}
Priyasad, D., Fernando, T., Denman, S., Sridharan, S., Fookes, C.: Attention driven fusion for multi-modal emotion recognition. In: ICASSP (2020)

\bibitem{clip}
Radford, A., Kim, J.W., Hallacy, C., Ramesh, A., Goh, G., Agarwal, S., Sastry, G., Askell, A., Mishkin, P., Clark, J., et~al.: Learning transferable visual models from natural language supervision. ICML  (2021)

\bibitem{radford2022whisper}
Radford, A., Kim, J.W., Xu, T., Brockman, G., McLeavey, C., Sutskever, I.: Robust speech recognition via large-scale weak supervision. arXiv preprint arXiv:2212.04356  (2022)

\bibitem{mag_transformer}
Rahman, W., Hasan, M.K., Lee, S., Bagher~Zadeh, A., Mao, C., Morency, L.P., Hoque, E.: Integrating multimodal information in large pretrained transformers. In: ACL (2020)

\bibitem{sarcasm_2}
Rockwell, P.: Lower, slower, louder: Vocal cues of sarcasm. Journal of Psycholinguistic research  (2000)

\bibitem{demucs_last}
Rouard, S., Massa, F., D{\'e}fossez, A.: Hybrid transformers for music source separation. In: ICASSP (2023)

\bibitem{avlnet}
Rouditchenko, A., Boggust, A., et~al.: {AVLnet: Learning Audio-Visual Language Representations from Instructional Videos}. In: INTERSPEECH (2021)

\bibitem{old_laughter_detector}
Ryokai, K., Dur\'{a}n~L\'{o}pez, E., Howell, N., Gillick, J., Bamman, D.: Capturing, representing, and interacting with laughter. In: Conference on Human Factors in Computing Systems (2018)

\bibitem{sablayrolles2019spreading}
Sablayrolles, A., Douze, M., Schmid, C., J{\'e}gou, H.: Spreading vectors for similarity search. In: ICLR (2019)

\bibitem{cola}
Saeed, A., Grangier, D., Zeghidour, N.: Contrastive learning of general-purpose audio representations. In: ICASSP (2021)

\bibitem{facenet}
Schroff, F., Kalenichenko, D., Philbin, J.: Facenet: A unified embedding for face recognition and clustering. In: CVPR (2015)

\bibitem{audio_vision_3}
Senocak, A., Oh, T.H., Kim, J., Yang, M.H., Kweon, I.S.: Learning to localize sound source in visual scenes. In: CVPR (2018)

\bibitem{shen2023fine}
Shen, X., Li, D., Zhou, J., Qin, Z., He, B., Han, X., Li, A., Dai, Y., Kong, L., Wang, M., et~al.: Fine-grained audible video description. In: CVPR (2023)

\bibitem{laughlog}
Shimasaki, A., Ueoka, R.: Laugh log: E-textile bellyband interface for laugh logging. In: Conference Extended Abstracts on Human Factors in Computing Systems (2017)

\bibitem{singer2022make}
Singer, U., Polyak, A., Hayes, T., Yin, X., An, J., Zhang, S., Hu, Q., Yang, H., Ashual, O., Gafni, O., et~al.: Make-a-video: Text-to-video generation without text-video data. arXiv preprint arXiv:2209.14792  (2022)

\bibitem{gpt2}
Solaiman, I., Brundage, M., Clark, J., Askell, A., Herbert{-}Voss, A., Wu, J., Radford, A., Wang, J.: Release strategies and the social impacts of language models. CoRR  (2019)

\bibitem{tanCOMMA2021}
Tan, R., Plummer, B.A., Saenko, K., Jin, H., Russell, B.: Look at what i'm doing: Self-supervised spatial grounding of narrations in instructional videos. In: NeurIPS (2021)

\bibitem{sarcasm_3}
Tepperman, J., Traum, D., Narayanan, S.S.: `yeah right': Sarcasm recognition for spoken dialogue systems. In: INTERSPEECH (2006)

\bibitem{audio_vision_4}
Tian, Y., Shi, J., Li, B., Duan, Z., Xu, C.: Audio-visual event localization in unconstrained videos. In: ECCV (2018)

\bibitem{tong2022videomae}
Tong, Z., Song, Y., Wang, J., Wang, L.: Videomae: Masked autoencoders are data-efficient learners for self-supervised video pre-training. NeurIPS  (2022)

\bibitem{torralba2011unbiased}
Torralba, A., Efros, A.A.: Unbiased look at dataset bias. In: CVPR (2011)

\bibitem{touvron2023llama}
Touvron, H., Lavril, T., Izacard, G., Martinet, X., Lachaux, M.A., Lacroix, T., Rozière, B., Goyal, N., Hambro, E., Azhar, F., Rodriguez, A., Joulin, A., Grave, E., Lample, G.: Llama: Open and efficient foundation language models. arXiv preprint arXiv:2302.13971  (2023)

\bibitem{llama2}
Touvron, H., Martin, L., Stone, K., Albert, P., et~al: Llama 2: Open foundation and fine-tuned chat models. arXiv preprint arXiv:2307.09288  (2023)

\bibitem{transformer}
Vaswani, A., Shazeer, N., Parmar, N., Uszkoreit, J., Jones, L., Gomez, A.N., Kaiser, {\L}., Polosukhin, I.: Attention is all you need. In: NeurIPS (2017)

\bibitem{wang2021multimodal}
Wang, L., Luc, P., Recasens, A., Alayrac, J.B., Oord, A.v.d.: Multimodal self-supervised learning of general audio representations. arXiv preprint arXiv:2104.12807  (2021)

\bibitem{wang2020event}
Wang, T., Zheng, H., Yu, M., Tian, Q., Hu, H.: Event-centric hierarchical representation for dense video captioning. IEEE Transactions on Circuits and Systems for Video Technology  (2020)

\bibitem{cross_attn_2}
Wei, X., Zhang, T., Li, Y., Zhang, Y., Wu, F.: Multi-modality cross attention network for image and sentence matching. In: CVPR (2020)

\bibitem{rjokes}
Weller, O., Seppi, K.: The rjokes dataset: a large scale humor collection. In: LREC (2020)

\bibitem{wav2clip}
Wu, H.H., Seetharaman, P., Kumar, K., Bello, J.P.: Wav2clip: Learning robust audio representations from clip. arXiv preprint arXiv:2110.11499  (2021)

\bibitem{xin2023improving}
Xin, Y., Yang, D., Zou, Y.: Improving text-audio retrieval by text-aware attention pooling and prior matrix revised loss. In: ICASSP (2023)

\bibitem{xue2023clip}
Xue, H., Sun, Y., Liu, B., Fu, J., Song, R., Li, H., Luo, J.: Clip-vip: Adapting pre-trained image-text model to video-language representation alignment. ICLR  (2023)

\bibitem{yang2023vid2seq}
Yang, A., Nagrani, A., Seo, P.H., Miech, A., Pont-Tuset, J., Laptev, I., Sivic, J., Schmid, C.: Vid2seq: Large-scale pretraining of a visual language model for dense video captioning. In: CVPR (2023)

\bibitem{yoon2018multimodal}
Yoon, S., Byun, S., Jung, K.: Multimodal speech emotion recognition using audio and text. In: IEEE Spoken Language Technology workshop (2018)

\bibitem{accent}
Zadeh, A., Liang, P.P., Mazumder, N., Poria, S., Cambria, E., Morency, L.P.: Memory fusion network for multi-view sequential learning. In: AAAI (2018)

\bibitem{audio_vision_10}
Zhou, H., Xu, X., Lin, D., Wang, X., Liu, Z.: Sep-stereo: Visually guided stereophonic audio generation by associating source separation. In: ECCV (2020)

\bibitem{zhu2022end}
Zhu, W., Pang, B., Thapliyal, A.V., Wang, W.Y., Soricut, R.: End-to-end dense video captioning as sequence generation. In: ACL (2022)

\end{thebibliography}

\end{document}